\documentclass[preprint,12pt]{elsarticle}

\usepackage{amssymb}
\usepackage{amsmath}

\newcommand \name{HPG-Diff}
\usepackage{subcaption}
\usepackage{booktabs}
\usepackage{amsmath}
\usepackage{xcolor}
\usepackage{array}
\usepackage{algorithm}
\usepackage{multirow}

\newcommand{\revise}[1]{\textcolor{black}{#1}}

\journal{Applied Soft Computing}

\begin{document}

\begin{frontmatter}



\title{\name{}: Hierarchical physics-guided diffusion with differentiable connectivity constraints for topology optimization} 

%

\author[label1]{Jinbo Yang\fnref{myfootnote}} 
\author[label2]{Mingyue Yuan\fnref{myfootnote}}
\author[label3]{Boyuan Zhang}
\author[label4]{Yoshifumi Kitamura}
\author[label1]{Shikai Jing \corref{cor1}}

\cortext[cor1]{Corresponding author. E-mail address: jingshikai@bit.edu.cn (S. Jing)}
\fntext[myfootnote]{These authors contributed equally to this work.}

\affiliation[label1]{organization={School of Mechanical Engineering, Beijing Institute of Technology},
            city={Beijing},
            postcode={100081}, 
            country={China}}

\affiliation[label2]{organization={Computer Science and Engineering, University of New South Wales},
            city={Sydney},
            postcode={2052},
            country={Australia}}

\affiliation[label3]{organization={College of Intelligence and Computing, Tianjin University},
            city={Tianjin},
            postcode={300072},
            country={China}}

\affiliation[label4]{organization={Research Institute of Electrical Communication, Tohoku University},
            city={Sendai},
            postcode={980-8576},
            country={Japan}}

\begin{abstract}
Deep generative models offer a promising paradigm for topology optimization, enabling rapid design exploration. 
However, these approaches lack intrinsic physics guidance, often leading to poor generalizability across unseen boundary conditions and the formation of floating material artifacts. 
To address these limitations, we propose Hierarchical Physics-Guided Diffusion (HPG-Diff), a novel diffusion framework that enforces physics consistency through two synergistic mechanisms.
First, we introduce a hierarchical physics-guided strategy that aligns different precomputed physics features with the denoising process, guiding material distribution toward optimal load paths to enhance generalizability. 
Second, we propose a floating material suppression loss as a differentiable connectivity constraint inspired by thermal conduction to improve topological connectivity.
By simulating a virtual heat propagation process from load positions, this mechanism explicitly penalizes floating material during training.
\revise{Quantitative evaluations demonstrate that HPG-Diff achieves average compliance errors of 0.87\% (in-distribution) and 5.29\% (out-of-distribution), while reducing floating material ratios to 2.90\% and 2.44\%, respectively.}
\revise{Furthermore, case studies on a 3:1 rectangular domain, including cantilever and bridge benchmarks, provide preliminary evidence that lightweight LoRA fine-tuning with a small dataset can support the adaptation of HPG-Diff to rectangular non-square domains.}
\end{abstract}





\end{frontmatter}



\section{Introduction}
Topology optimization (TO) has been widely used in various engineering design scenarios due to its capacity to generate lightweight structures with high mechanical performance. To date, numerous TO designed structures have been successfully applied in the aerospace \cite{zhu2016topology, dagkolu2021design, guanghui2020aerospace}, mechanical \cite{andersen2021competition, ferrari2021topology, maestre2023toros}, civil \cite{bi2022topology, yang2023integrated}, and biomedical \cite{ichihara20223d, garcia2024dynamic, li2023compressive} fields. This iterative optimization approach typically relies on Finite Element Analysis (FEA). A common example is the Solid Isotropic Material with Penalization (SIMP)~\cite{bendsoe1989optimal}. However, it is computationally demanding, especially in large‑scale optimization problems.

\begin{figure}[htbp]
  \includegraphics[width=\columnwidth]{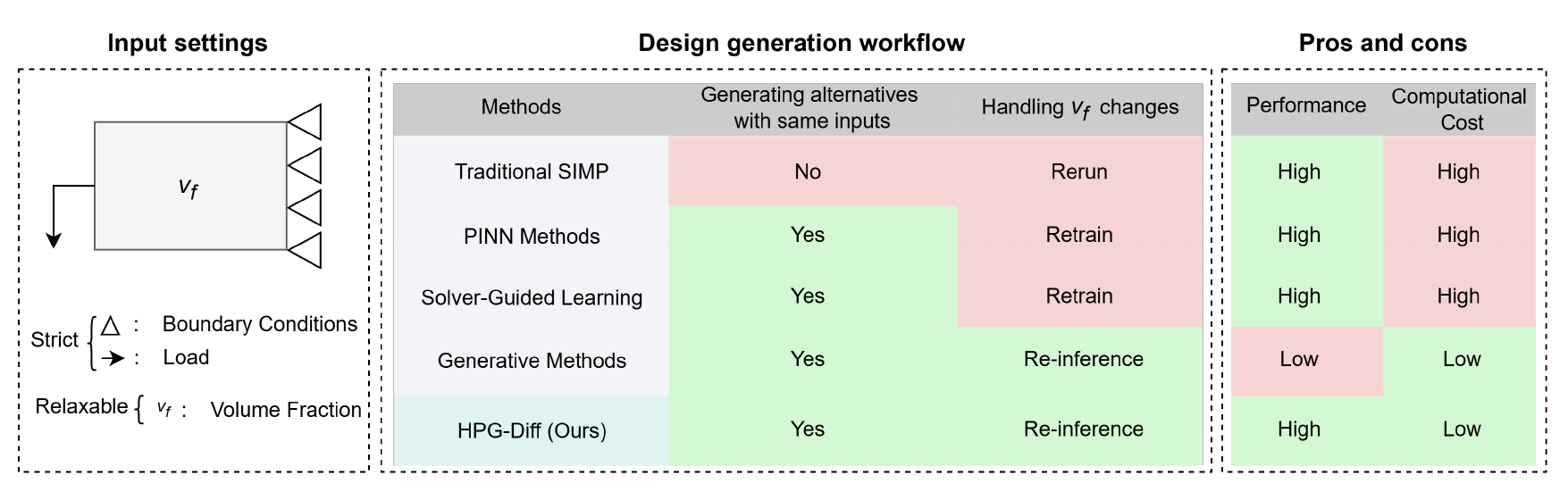} 
  \caption{\textbf{Performance vs Efficiency: Engineering design demands multiple high-performing alternatives.} Comparison of mainstream TO methods in adaptability, performance, and computational cost. For a given TO design, strict inputs (boundary conditions, load) must always be satisfied, while relaxable inputs (volume fraction) allow adjustment, resulting in two modes of generating new designs.}
  \label{fig:intro}
\end{figure}

\revise{Data-driven design methods have also been widely explored in broader engineering domains~\cite{seyed2023design,xu2026database,bu2025data,chen2025performance,wang2024reverse}.}
In topology optimization, deep learning has significantly reduced computational costs using surrogate models~\cite{zheng2021data, kallioras2021dl, qian2023topology, yan2023real, yan2022deep}.
However, they often suffer from a lack of \textbf{physics consistency}: the mechanical performance and manufacturability may violate the physics principles. Since surrogate models approximate complex behavior by fitting input–output relationships, they may not effectively capture the physics representation of optimization objectives and constraints, potentially resulting in generated structures with anomalous stress concentrations or connectivity defects. 
\revise{For example, density-field surrogate predictions may require additional physics-based correction or post-processing when the learned mapping does not fully preserve equilibrium, volume, or connectivity-related constraints~\cite{deng2022self,luo2021improved}.}
At the same time, practical engineering design demands a diverse set of high-performing alternatives. As illustrated in Fig.~\ref{fig:intro}, this requires generating new designs, either by changing strict inputs for a fixed problem or by adjusting relaxable inputs to explore the design space. 
\revise{Strict inputs denote problem-defining conditions such as boundary conditions and loads, while relaxable inputs denote design-control variables such as volume fraction. The comparison in Fig.~\ref{fig:intro} is qualitative and used to illustrate typical design methods.}
However, mainstream TO methods face clear limitations when acquiring new designs. Traditional SIMP is deterministic, offering no diversity under same inputs, and requires a complete rerun for modified arguments. For methods like PINN\cite{jeong2023physics,jeong2025advanced} and solver-guided learning~\cite{deng2022self, chandrasekhar2021tounn}, both generating diversity for a fixed problem by changing random seeds and adapting to input changes require same time-consuming retraining process. Consequently, designers have to choose between efficiency and quality, highlighting the need for a TO framework that enables fast, diverse, and high-performing design generation.

\revise{
This trade-off is not only computational but also reflects different ways of enforcing physics constraints. 
PINN and solver-guided methods can impose relatively better physics consistency by embedding governing-equation residuals, energy terms, or online FEA feedback into optimization. 
However, this constraint hardness often reduces generative flexibility, because new boundary conditions, loads, or design requirements may require per-case optimization, retraining, or iterative solver interaction. 
Generative models provide a complementary route: after pretraining, they can efficiently sample diverse candidate designs, but their physics consistency depends on the design of conditioning signals and training objectives. 
}

Recently, advances in Generative Adversarial Network (GAN)~\cite{goodfellow2014generative} and diffusion models~\cite{ho2020denoising} have improved generation quality through more precise modeling of data distributions.
Meanwhile, these generative methods inherently have the capacity for fast and diverse generation.
For TO, some physics-guided variants \cite{regenwetter2022deep} further integrate physics knowledge to improve performance.

While physics knowledge has been incorporated into these methods~\cite{maze2023diffusion, giannone2023aligning, zhang2025research}, several challenges remain in practical applications. The reasons are twofold:
(1) These methods treat physics knowledge as either global inputs or external steering signals, overlooking the intrinsic correlation between hierarchical nature of the denoising process and distinct physical knowledge. Consequently, mechanical fidelity is limited due to the lack of generalizability, especially in out-of-distribution scenarios
(2) Standard pixel-wise objectives are inherently topology-agnostic. Despite improvements, generative models still produce a higher proportion of floating material than traditional SIMP-like methods.

\begin{figure}[htbp]
  \includegraphics[width=\columnwidth]{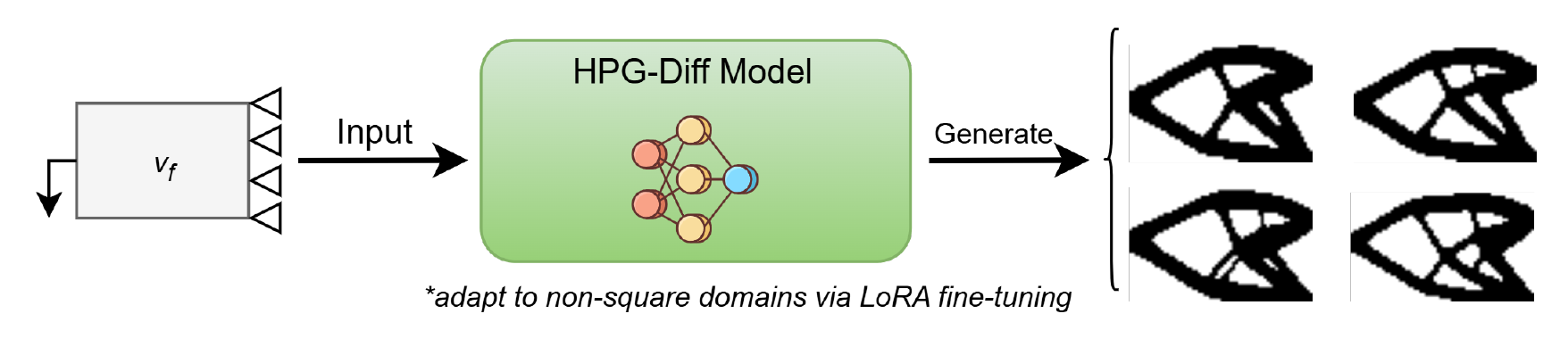} 
  \caption{\textbf{Example of \name{} for rapid and diverse topology optimization design generation.} Given the boundary conditions and volume fraction, the model generates diverse and high-performing structures. Note that LoRA fine-tuning is employed to adapt the model to non-square design domains.}
  \label{fig:workflow}
\end{figure}

This paper aims to address these challenges to develop an innovative method that improves mechanical fidelity and manufacturability in TO, while retaining the efficiency and diversity of generative methods. In order to replicate both the optimal material distribution goal and the connectivity constraint of classical topology optimization (e.g., SIMP) within a generative paradigm, we propose \textbf{Hierarchical Physics-Guided Diffusion (\name{})}, a diffusion framework that rapidly generates diverse high-performing designs (Fig.~\ref{fig:workflow}) through the synergistic integration of hierarchical feature guidance and differentiable connectivity constraints.

Specifically, we introduce a hierarchical physics-guided strategy to guide the denoising process. By incorporating three key mechanical responses \textit{Displacement (U)}, \textit{Principal Stress Line (PSL)}, and \textit{Strain Energy Density (SED)}, we capture deformation magnitude, critical load paths, and stiffness-contributing regions, respectively. Building upon approaches that rely primarily on high-level features (e.g., \textit{SED})~\cite{maze2023diffusion}, we introduce low-level \textit{U} and mid-level \textit{PSL} as novel contexts hierarchically injected across network layers. This aligns the model's receptive fields with distinct physics features, ensuring the optimal material distribution. Moreover, we impose a floating material suppression loss as a differentiable connectivity constraint to regularize the output topology. This thermal-conduction-inspired mechanism ensures the material forms a continuous structure by explicitly penalizing disconnected regions during training, which effectively reduces floating material artifacts in the generated structures.
Our main contributions are:

\noindent 
\begin{itemize} 
    \item \revise{A hierarchical physics-guidance strategy is proposed for topology optimization, which organizes distinct key physics features as layer-specific conditioning signals aligned with different stages of the denoising process. This strategy enhances mechanical fidelity and generalizability to unseen boundary conditions.}
    \item The discrete topological connectivity problem is reformulated as a differentiable propagation process to derive a floating material suppression loss. This allows the connectivity constraint to be explicitly integrated into the gradient-based training loop in generative TO.
    \item Our diffusion framework \name{} balances high mechanical performance with the generation efficiency. Notably, case studies demonstrate that the learned physics priors enable rapid adaptation to different design domains via lightweight LoRA fine-tuning.
\end{itemize}

\section{Related works}

\subsection{Physics-aware deep learning paradigms in topology optimization}

\revise{Learning-based computational methods have been increasingly explored across constrained engineering domains, such as infrastructure assessment and energy-system operation~\cite{bayat2026deep, li2025temporally, zeng2026deriving, feng2025aggregation}.}
Topology optimization (TO) aims to compute optimal material distribution under given loads and boundary conditions, and it is widely used in industrial design. In real-world scenarios, designers require not only high-performing structures but also multiple alternative solutions to trade off manufacturability, cost, and aesthetics.

Classical numerical methods such as SIMP~\cite{bendsoe1989optimal} provide physically consistent results but are computationally expensive for diverse design explorations. To address this challenge, deep learning frameworks have emerged as promising alternatives~\cite{qiu2021deep, kim2021machine, brown2022deep, white2019multiscale}.

Recent work has emphasized integrating physics knowledge into learning pipelines. We roughly categorize existing approaches into three paradigms. \textbf{PINN methods} (e.g., PINNTO~\cite{jeong2023physics} and CPINNTO~\cite{jeong2025advanced}) embed governing equations or physics terms in the loss. \textbf{Solver-guided learning} methods (e.g., SOLO~\cite{deng2022self} and TOuNN~\cite{chandrasekhar2021tounn}) incorporate online FEA feedback during training. These two paradigms improve mechanical performance, but typically require per-case retraining or iterative FEA updates when the problem setting changes, which limits scalability in design-exploration workflows.

In contrast, \textbf{generative methods} (GANs, diffusion models)~\cite{nie2021topologygan, maze2023diffusion, giannone2023aligning, zhang2025research} emphasize rapid, diverse sample generation: after pretraining they can produce multiple candidate designs at inference without per-case retraining. Table~\ref{tab:one} summarizes a comparison of these physics-aware deep learning paradigms (per-case training, support for diverse outputs, and relative inference cost). 
\revise{The inference-time multipliers are approximate ranges. They are summarized from the typical behavior of mainstream implementations, and may vary with detailed implementations or settings.}

\begin{table}[htbp]
  \centering
  \footnotesize
  \begin{tabular}{l >{\centering\arraybackslash}p{2cm} >{\centering\arraybackslash}p{2.7cm} >{\centering\arraybackslash}p{2.3cm}}
    \toprule
    \textbf{Method} & \textbf{PINN Methods} & \textbf{Solver-Guided Learning} & \textbf{Generative Methods} \\
    \midrule
    \textbf{Per-Case Training Required} & Yes & Yes & No \\
    \textbf{Supports Diverse Outputs} & No & No & Yes \\
    \textbf{Inference Time} & 50$\sim$100X & 2$\sim$3X & X \\
    \bottomrule
  \end{tabular}
  \caption{Comparison of physics-integration strategies for mainstream deep learning methods. The method classifications reflect typical mainstream implementations, and inference time is reported as an approximate multiple of baseline Generative Methods (X).}
  \label{tab:one}
\end{table}

\subsection{Physics-conditioning mechanisms in generative TO}

The effectiveness of generative models for TO hinges on how physics guidance is integrated—a process known as conditioning. Current strategies for diffusion models fall into two main categories:

\textbf{Architectural Modification} approaches offer fine-grained control by altering the model's structure, such as modifying the network backbone~\cite{zhang2023adding, mou2024t2i, zeng2024dilightnet}, adding new trainable modules~\cite{li2021sg}, or using internal attention mechanisms~\cite{li2023gligen}. \textbf{Sampling-Phase} methods steer the generation process without changing the model, typically by using external classifiers~\cite{dhariwal2021diffusion} or adapting the sampling algorithm~\cite{ho2022classifier, voynov2023sketch}.

In TO, diffusion model-based methods~\cite{maze2023diffusion, giannone2023aligning, zhang2025research} belong to the second. This often involves using physics features like \textit{Strain Energy Density (SED)} and \textit{von Mises stress} to represent input settings. While this strategy avoids issues with sparse data, such high-level features cannot fully capture the complex internal mechanics of a structure. Therefore, they often require complex multi-stage framework or auxiliary models, which actually lower the generation efficiency.

\revise{From the perspective of physics features integration, existing diffusion-based TO methods mainly introduce physics information as an overall condition or steering signal for the generation process~\cite{maze2023diffusion,giannone2023aligning,zhang2025research}. 
Such designs provide useful physical context, but they do not explicitly organize different physics features according to their distinct structural roles in TO. }
In this work, we implement an architecture-level, multi-level conditioning strategy that hierarchically aligns multi-level physics features with the denoising process, thereby providing richer mechanical context during generation while keeping an efficient pipeline.

\subsection{Topological connectivity and design domain adaptation}

For practical application, two issues remain challenging in generative TO. First, generated structures often contain disconnected or floating material that requires additional processing. Prior works typically rely on post-processing filters~\cite{sigmund2013topology, deaton2014survey, cool2025practical}, auxiliary classifiers~\cite{maze2023diffusion, patel2022improving}, or extra loss terms~\cite{luo2021improved, mahdi2021gantl} to mitigate these defects. 

Second, generative TO studies train and validate on square domains, whereas industrial components often design under non-square domains. This domain mismatch often requires retraining or substantial fine-tuning on target-specific datasets~\cite{zhang2025research}. To address domain adaptation with limited data, parameter-efficient fine-tuning methods (e.g. LoRA~\cite{hu2022lora}) can adapt a pretrained generative prior to new domain sizes or aspect ratios with small datasets. However, LoRA’s success depends on the generalizability of pretrained diffusion model and the similarity between pretraining and target data distributions. 

\revise{
In our work, we target these two problems in generative TO.
For connectivity, existing differentiable connectivity losses in shape generation typically act on the generated occupancy, mask, or implicit field to reduce broken or separate components and improve object-level connectivity~\cite{mosinska2018beyond,hu2019topology,shit2021cldice}. This objective is largely geometry-level: it measures whether the generated shape is connected as an object. In contrast, our FMS addresses a topology-optimization-specific connectivity issue. It evaluates whether high-density material is reachable through the predicted material layout. Thus, the penalty is a load-seeded reachability penalty for TO.
For domain adaptation, }we evaluate LoRA-based adaptation in representative non-square cases in Section~\ref{sec:case}. Evaluation focuses on compliance error, connectivity, and design diversity to quantify applicability.

\section{Method}
\label{sec:method}

In this section, we demonstrate the structure of \name{}, which enables a single diffusion model to generate topological structures with high physics consistency. The section is organized as follows: Section~\ref{subsec:3.1} illustrates how we select physics information as features. Section~\ref{subsec:3.2} introduces the details of \name{} and the hierarchical physics-guided strategy. Section~\ref{subsec:3.3} depicts the loss function designed to control floating material artifacts.

\begin{figure*}[htbp] 
  \centering
  \includegraphics[width=0.7\columnwidth]{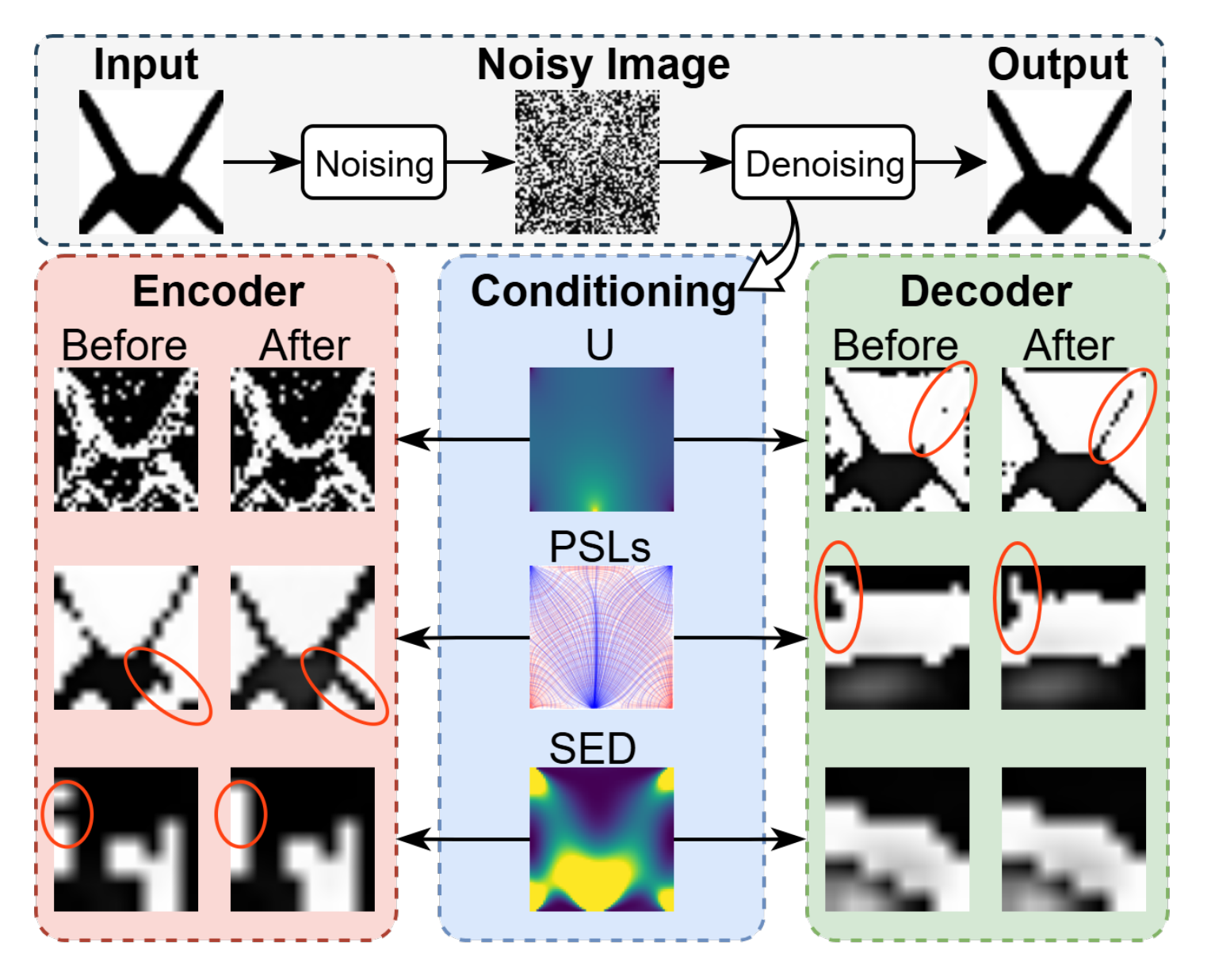}
  \caption{Example of local feature enhancements driven by conditioning physics features in the denoising process.
  Notable enhancements can be observed in the red-circled regions as a result of this conditioning.}
  \label{fig:one}
\end{figure*}

\subsection{Representation TO with physics features}
\label{subsec:3.1}

The minimum-compliance problem in TO can be written as:
\begin{equation}
\begin{aligned}
& \min_{x} & & c(x) = F^TU=U^T K U \\
& \text{subject to:} & & \frac{V(x)}{V_0} = v_f \\
& & & 0 < x_{\min} \le x \le 1
\end{aligned}
\end{equation}

where $U$ and $F$ are the global displacement and force vectors, respectively, $K$ is the global stiffness matrix, $x$ is the design variables, $x_{\min}$ is the minimum relative density, $V({x})$ and $V_0$ are the material volume and design domain volume, respectively. $v_f$ is the prescribed volume fraction.

The total potential energy of the structure is the difference between strain energy $E_{strain}$ and potential energy of the external loads $E_{loads}$. In discrete form, this is:
\begin{equation}
    \Pi = E_{strain} - E_{loads} = \frac{1}{2}{U}^T{K}{U} - {U}^T{F}
\end{equation}

According to the Principle of Minimum Potential Energy, the equilibrium state is found when $\frac{\partial \Pi}{\partial {U}} = 0$, which yields the system equation: 
\begin{equation}
\label{eq:three}
{K}{U} = {F}
\end{equation}

Solving Eq.~\ref{eq:three} via FEA yields the displacement field \textit{U}. 
While \textit{U} is the most fundamental low-level feature representing displacement of all elements, we derive two higher-level features from it to provide more direct insight into the optimized structure.

The first, \textbf{Strain Energy Density (SED)}, directly relates to the optimization objective. As compliance is twice the total strain energy ($c({x})={U}^T{K}{U}=2E_{strain}$), \textit{SED} pinpoints regions most critical to the structure's stiffness.

The second, \textbf{Principal Stress Line (PSL)} explains the structural morphology. Derived from the stress field, \textit{PSL} visualizes the internal load paths, thereby revealing why the topology forms its specific shape.

Fig.~\ref{fig:one} provides an illustrative example of the impact of physics features on conditioning the denoising process. The red-circled regions highlight how these features enhance the UNet's feature maps, improving the preservation of both details and overall structural integrity.

Specifically, the corresponding conditioning physics features are:
\textbf{(1) Displacement (U)}: Low-level feature that quantifies the element displacements within the design domain. Yellow and dark blue denote maximum and minimum displacements at the load and support points. The circled region shows the preserved thin rod structure, which is a critical load path.
\textbf{(2) Principal Stress Line (PSL)}: Mid-level feature that visualizes internal load paths via tensile (red) and compressive (blue) stresses. The circles highlight the encoder preserves a thin rod structure and the decoder improves overall structural connectivity.
\textbf{(3) Strain Energy Density (SED)}: High-level feature that indicates critical regions (yellow areas) related to stiffness. The circled region shows the encoder removes a redundant feature, thus enhancing structural integrity.

\begin{figure*}
\centering 
\includegraphics[width=\textwidth]{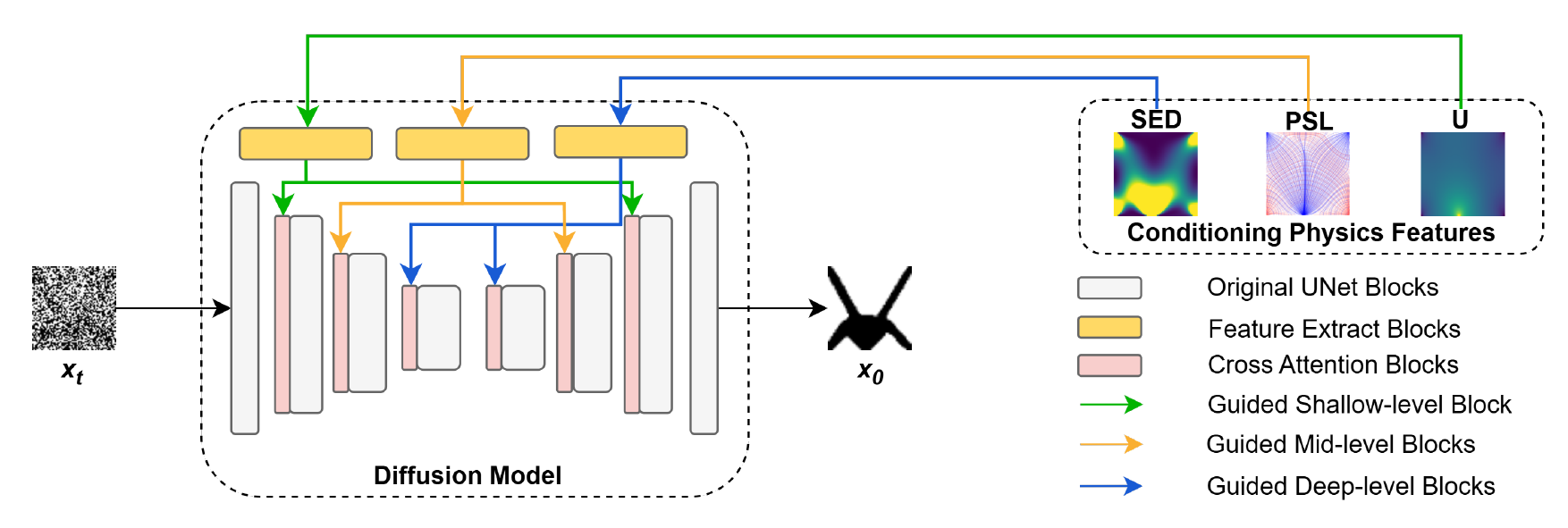}
  \caption{\textbf{(\name{}) Hierarchical Physics-Guided Diffusion Framework.} Colored lines depict how \textit{U} (green), \textit{PSL} (yellow), and \textit{SED} (blue) hierarchically guide shallow, mid, and deep layers via cross-attention modules. 
  }
  \label{fig:two}
\end{figure*}

\subsection{Hierarchical physics-guided strategy}
\label{subsec:3.2}

Diffusion models with standard UNet architectures inherently process information hierarchically. Through successive downsampling operations in the encoder path, the effective receptive field of convolutional kernels progressively increases. As a result, deeper layers capture more abstract, global structural context from the input. Shallower layers work differently: they use smaller receptive fields and operate closer to the original input resolution. This allows them to primarily focus on preserving and refining fine-grained local details.

Crucially, this hierarchical nature is mirrored in the decoder path. As the network upsamples the latent features back to the pixel space, the denoising process progressively reconstructs the topology. In the context of topology optimization, this implies a functional separation: transitioning from determining the global material layout (deep layers), through developing intermediate structural components (mid layers), to finish the details of structural boundaries (shallow layers).

We leverage this inherent architectural hierarchy by drawing a parallel to the distinct physics characteristics represented by our chosen physics features (Section \ref{subsec:3.1}). Specifically, \textit{Displacement (U)}, which describes point-wise deformation and local geometric changes, aligns naturally with the fine-grained and local feature processing capabilities of the shallower layers in the UNet. Conversely, \textit{Principal Stress Line (PSL)}, depicting the global pathways of load transmission, and the \textit{Strain Energy Density (SED)}, highlighting structurally significant regions of energy concentration, represent more global mechanical characteristics. These align conceptually with the global structural context captured effectively by the deeper layers of the UNet.

Based on this correspondence, we propose a hierarchical physics-guided strategy. The core principle is to condition the diffusion model's denoising process by injecting these distinct physics features at specific, corresponding stages or layers within the UNet architecture, rather than providing all conditioning information uniformly. This means guiding the generation process with the type of physics features that best matches the abstraction level and receptive field size of the UNet features at that particular stage. As Fig.~\ref{fig:two} shows, we employ a layer-specific cross-attention mechanism within the UNet's encoder and decoder. At designated layers (e.g., shallow, mid, deep layers), cross-attention modules are introduced to dynamically attend to the feature representations of the corresponding injected physics features. The denoising process is illustrated as:

\begin{equation}
\label{eqn:denoise}
p_\theta(x_{t-1} \mid x_t, c) = \mathcal{N}\left(x_{t-1} \mid \mu_\theta(x_t, t, c), \Sigma_\theta(t)\right)
\end{equation}
where $c$ represents the physics guidance condition.

This hierarchical injection combined with the cross-attention mechanism facilitates a dynamic coupling between the generated topology and the mechanical principles represented by the physics features. 
Physics consistency is better maintained during the sampling process by aligning the physics guidance with the appropriate layers of the UNet, leading to generated topologies that are more physically plausible.

\subsection{Floating material suppression loss}
\label{subsec:3.3}

Our proposed framework aims to replicate the rigorous physical consistency of classical topology optimization (e.g., SIMP) within a generative paradigm. In SIMP, FEA-driven sensitivity analysis serves dual purposes: it simultaneously guides material distribution to minimize compliance and inherently enforces connectivity, since disconnected elements carry no load and are eliminated by the optimization process.

In contrast, standard diffusion models lack this intrinsic physical feedback. While the hierarchical physics-guided strategy (Section~\ref{subsec:3.2}) effectively directs material allocation by conditioning on physics fields, it functions primarily as a soft guidance mechanism rather than a strict constraint. Meanwhile, the generative process is driven by pixel-wise denoising objectives (e.g., $L_2$ MSE). These losses treat pixels as statistically independent variables and are inherently topology-agnostic, making the model prone to generating floating material, such as isolated islands that degrade mechanical performance and manufacturability.

\revise{To mitigate this issue, we propose a loss function inspired by the physical mechanism of thermal conduction~\cite{crane2013geodesics}, based on the intuition that heat-flow propagation from a prescribed source provides a continuous way to measure reachability through connected material paths.}
Unlike standard discrete connectivity checks (e.g., Connected Component Labeling) which are non-differentiable, we formulate connectivity as a differentiable heat propagation process.

The core logic operates by placing a virtual heat source at the load position and simulating its propagation through the predicted material. Effectively, any material that remains "cold" implies it is structurally disconnected. This mechanism is implemented through three coupled stages:

\begin{enumerate}
    \item \textbf{Seeding:} Initialize the propagation source with the binary seed mask $\mathbf{S}_{\mathrm{load}}\in\{0,1\}^{N\times N}$ at the load position. Since the load position serves as the functional origin of the stress transmission path, only material connected to this source is recognized as part of the effective structure.
    \item \textbf{Differentiable Propagation:} Allow a signal to iteratively propagate over the predicted material distribution $\hat{\mathbf{x}}_{\mathrm{pred}}$ to obtain the connectivity strength map $\mathbf{D}_{\mathrm{final}}$.
    \item \textbf{Penalty:} After propagation converges, penalize regions that exhibit high material density (approaching 1) but low connectivity signal (approaching 0). This penalty encourages the model to suppress these isolated components that fail to receive the propagated signal.
\end{enumerate}

$\mathbf{D}_{\mathrm{final}}$ is computed via the differentiable propagation. Let $\mathbf{S}_{\mathrm{load}}\in\{0,1\}^{N\times N}$ be a binary seed mask indicating load location. Define the local maximum operator (equivalent to a $3\times3$ dilation)

\begin{equation}
\big(\mathrm{Max}_{3\times3}(\mathbf{D})\big)_{ij}=\max_{(u,v)\in\mathcal{N}_1(i,j)}D_{uv}
\end{equation}
and initialize $\mathbf{D}^{(0)}=\mathbf{S}_{\mathrm{load}}$. We update
\begin{equation}
D_{ij}^{(k+1)} = \max\big( D_{ij}^{(k)}, (\mathrm{Max}_{3\times3}(\mathbf{D}^{(k)}))_{ij} \, (\hat{\mathbf{x}}_{\text{pred}})_{ij} \big)
\end{equation}
and we iterate this process for $k$ steps (where $k$ covers the domain dimension) $\mathbf{D}_{\mathrm{final}}=\mathbf{D}^{(k)}$.

\revise{
This propagation provides a justification for FMS in topology optimization.
In the binary-density limit, the predicted density field defines a solid mask $\Omega_s$, and the load mask $S_{\mathrm{load}}$ defines the source set.
The max-propagation step can be interpreted as a masked reachability dilation: the connectivity signal is allowed to expand only through elements belonging to $\Omega_s$.
Equivalently, the discrete reachable set follows $\mathcal{R}^{(0)} = S_{\mathrm{load}}$,
$\mathcal{R}^{(k+1)} = \mathcal{R}^{(k)}
\cup
\{i\in\Omega_s \mid i \text{ is adjacent to some } j\in\mathcal{R}^{(k)}\}$.
By induction, $\mathcal{R}^{(k)}$ contains the solid elements that can be reached from the load seed through a continuous solid path of at most $k$ steps.
Therefore, when $k$ covers the domain scale, the propagation converges to the reachable solid component.
For continuous density predictions, the final connectivity map $\mathbf{D}_{\mathrm{final}}$ can be viewed as a continuous approximation of this binary reachable component, where larger values indicate stronger reachability from the load seed through the predicted material field.
Thus, high-density regions with low propagation values correspond to floating material that is not incorporated into the load-seeded material component.
This is consistent with topology optimization, where disconnected solid regions consume the prescribed volume budget but do not effectively contribute to the main load-transfer structure.
}

Thus we introduce the \textbf{Floating Material Suppression Loss} (FMS loss), denoted as $\mathcal{L}_{\mathrm{fm}}$, which penalizes floating material:

\begin{equation}
\label{eq:lfm}
\mathcal{L}_{\mathrm{fm}}
= \mathbb{E}_{\mathbf{x}_0,\epsilon,t}\!\left[ w(t)\;\frac{1}{|\Omega|}\sum_{(i,j)\in\Omega}(\hat{\mathbf{x}}_{\text{pred}})_{ij}\big(1-({\mathbf{D}_{\text{final}})_{ij}}\big) \right]
\end{equation}
where $\hat{\mathbf{x}}_{\mathrm{pred}}$ is the predicted material distribution, $\mathbf{D}_{\mathrm{final}}\in[0,1]^{N\times N}$ is the connectivity strength map obtained from the differentiable propagation, $t$ is the diffusion timestep, $T$ the maximal timestep, and $c>0$ a scalar controlling the time weighting.

We use an exponential time-weighting $w(t)=\exp(-c\,t/T)$ so the connectivity prior acts mainly at late denoising steps $(t \to 0)$, avoiding gradient conflict with $\mathcal{L}_{\mathrm{noise}}$. 
\revise{
This weighting improves the stability of FMS during early denoising stages. 
When $t$ is large, the predicted density field is still highly noisy and may not provide a coherent propagation medium; the small value of $w(t)$ prevents the FMS term from producing dominant gradients at this stage. 
As $t$ approaches 0, the predicted density field becomes more structured, and FMS becomes active on high-density regions that remain unreachable from the load seed. 
Because the penalty is multiplied by $\hat{\mathbf{x}}_{\mathrm{pred}}$, near-void regions contribute little to the loss, while disconnected high-density regions receive stronger gradients.
}
\revise{Therefore, the scalar $c$ is defined relative to $T$. We further provide sensitivity analysis in Appendix C.}
The final training objective combines $\mathcal{L}_{\mathrm{fm}}$, with $\lambda_{\mathrm{fm}}$ balancing denoising loss and topology regularization:
\begin{equation}
\mathcal{L} \;=\; \mathcal{L}_{\mathrm{noise}} \;+\; \lambda_{\mathrm{fm}} \,\mathcal{L}_{\mathrm{fm}}
\end{equation}

\section{Experiments}
\label{sec:four}

\subsection{Dataset}
\label{sec:Dataset}
We utilize the dataset from Mazé et al.~\cite{maze2023diffusion}, which was collected using the SIMP method and contains 30,000 training samples and two test sets. 
\textbf{Test 1} includes 1,800 samples with in-distribution boundary conditions, while \textbf{Test 2} contains 1,000 samples with out-of-distribution boundary conditions. 
The dataset was generated under varied conditions: the volume fraction ranges from 0.3 to 0.5 in 0.02 increments; loads are applied at random unconstrained boundary nodes with directions from $[0, \pi]$ in $\pi/6$ intervals; and 42 boundary conditions were used for training versus 5 new boundary conditions for testing.
Each sample in the dataset provides the optimal topology, minimum compliance, and details on loads, boundary conditions, and volume fraction. 
The original data also includes preprocessed \textit{von Mises stress} and \textit{strain energy density}. 
Moreover, we have supplemented this dataset with \textit{displacement} and \textit{principal stress line} as additional physics features.

\subsection{Evaluation metrics}
\label{sec:Metrics}
Distinct from common metrics used for generative models which typically assess the quality, realism, and diversity of generated images, we use physics-based metrics adapted from \cite{maze2023diffusion}. Given the primary objective of topology optimization (TO) is to achieve minimum compliance, the following three metrics are employed. For all metrics, evaluations are performed on binarized results (threshold $\tau=0.5$) using a standard FEA solver to ensure fair comparisons:

\textbf{Compliance Error (CE)} is the relative error in compliance compared to ground truth; 

\textbf{Volume Fraction Error (VFE)} is the relative error in volume fraction compared to ground truth; 

\textbf{Floating Material (FM)} is the proportion of samples containing floating materials.

These evaluation metrics are applied to both the \textbf{Test 1} and \textbf{Test 2}. The evaluation on the \textbf{Test 2} specifically aims to assess the model's generalizability on unseen data distributions, thereby indicating whether the model has effectively learned the optimization physics mechanisms.

\subsection{Implementation}
\label{sec:Implementation}
We implement our \name{} based on PyTorch using a single NVIDIA A100 for both training and inference. For training, we initiate the training process with diffusion steps 1000 for around 200 epochs. The AdamW \cite{loshchilov2017decoupled} optimizer was utilized along with a batch size set to 64. We implemented the cosine noise schedule with an initial learning rate of 0.0001.

\subsection{Baseline comparison}

\subsubsection{Setup}

To better evaluate \name{}, we compare \name{} with three SOTA generative methods (TopologyGAN \cite{nie2021topologygan}, TopoDiff \cite{maze2023diffusion}, and DOM \cite{giannone2023aligning}). Using the datasets and evaluation metrics mentioned in Section \ref{sec:Dataset} and \ref{sec:Metrics}.

\subsubsection{In-Distribution}

\begin{table}[htbp]
    \centering
    \begin{tabular}{lcccc}
        \toprule
        Method & Avg CE $\downarrow$ & Med CE $\downarrow$ & Avg VFE $\downarrow$ & FM $\downarrow$ \\
        \midrule
        TopologyGAN \cite{nie2021topologygan} & 48.51 & 2.06 & 11.87 & 46.78 \\
        TopoDiff-Guided \cite{maze2023diffusion} & 4.39 & 0.83 & 1.85 & 5.54 \\
        DOM w/ TA \cite{giannone2023aligning} & 4.44 & 0.74 & 1.52 & 6.72 \\
        \textbf{\name{}} (Ours) & \textbf{0.87} & \textbf{0.16} & \textbf{1.38} & \textbf{2.90} \\
        \bottomrule
    \end{tabular}
    \caption{\textbf{Baseline Comparison:} Performance (in \%) on the \textbf{Test 1 (in-distribution)} dataset.}
    \label{tab:two_1}
\end{table}

As Table~\ref{tab:two_1} shows, compared to three baseline methods, \name{} reduces the CE, achieving values of 0.87\% in average and 0.16\% in median. Furthermore, \name{} effectively suppresses the existence of FM to 2.90\%. average VFE is reduced to 1.38\%. 

When compared to the other generative methods, \name{} reduces the Average CE by 80.18\% relative to TopoDiff-Guided and by 80.41\% relative to DOM w/ TA. The FM is also reduced by 47.65\% and 56.85\% compared to TopoDiff-Guided and DOM w/ TA, respectively. Compared with TopologyGAN, our method improves Average CE, Median CE, and FM by approximately 98.21\%, 92.23\%, and 93.80\%.

\subsubsection{Out-of-Distribution}

\begin{table}[htbp]
    \centering
    \begin{tabular}{lcccc}
        \toprule
        Method & Avg CE $\downarrow$ & Med CE $\downarrow$ & Avg VFE $\downarrow$ & FM $\downarrow$ \\
        \midrule
        TopologyGAN \cite{nie2021topologygan} & 143.08 & 6.82 & 14.31 & 67.90 \\
        TopoDiff-Guided \cite{maze2023diffusion} & 18.40 & 1.82 & 1.80 & 6.21 \\
        DOM w/ TA \cite{giannone2023aligning} & 32.19 & 3.69 & 1.78 & 14.20 \\
        \textbf{\name{}} (Ours) & \textbf{5.29} & \textbf{0.61} & \textbf{1.45} & \textbf{2.44} \\
        \bottomrule
    \end{tabular}
    \caption{\textbf{Baseline Comparison:} Performance (in \%) on the \textbf{Test 2 (out-of-distribution)} dataset.}
    \label{tab:two_2}
\end{table}

Table~\ref{tab:two_2} shows that \name{} outperformed in Average CE, Median CE, and FM, achieving values of 5.29\%, 0.61\% and 2.44\%, respectively. average VFE is reduced to 1.45\%. 

Statistically, \name{} substantially outperforms all baselines in the out-of-distribution tests. It reduces the key Average CE metric by 71.25\% and 83.57\% compared to TopoDiff-Guided and DOM w/ TA, respectively, while also improves Median CE and FM by 66.48\%$\sim$83.47\% and 60.71\%$\sim$82.82\% against these two methods. When compared to TopologyGAN, all major errors in CE and FM are reduced by over 91\%. These results show the potential of generalizability of \name{} for unseen design problems.

Multi-stage methods often trade higher computational cost for performance gains. We investigate this by comparing our \name{} with DOM followed by SIMP post-processing on out-of-distribution tests (Table~\ref{tab:three}). While increasing the number of SIMP steps from 5 to 10 improves the baseline's scores, it does not surpass our method's median CE and FM. Furthermore, this improvement comes at the direct cost of increased inference time, as each SIMP step requires an extra FEA calculation. \name{} achieves competitive performance without this iterative overhead, demonstrating the effectiveness of our framework.

\begin{table}[htbp]
    \centering
    \begin{tabular}{lccc}
        \toprule
        Method & Mdn CE (\%) $\downarrow$ & Avg VFE (\%) $\downarrow$ & FM (\%) $\downarrow$ \\
        \midrule
        DOM w/ TA & 3.69 & 1.78 & 14.20 \\
        DOM w/ TA + SIMP$_5$ & 1.89 & 1.77 & 10.19 \\
        DOM w/ TA + SIMP$_{10}$ & 1.15 & 1.10 & 2.61 \\
        \textbf{\name{}} & \textbf{0.61} & 1.45 & \textbf{2.44} \\
        \bottomrule
    \end{tabular}
    \caption{Comparison with multi-stage method  in \textbf{out-of-distribution} tests. Subscript of SIMP denotes the number of iteration steps of SIMP.}
    \label{tab:three}
\end{table}

\subsection{Ablation study} 
\label{sub:abl}

Table~\ref{tab:four_1} and Table~\ref{tab:four_2} show the ablations on Hierarchical Physics-Guided (HPG) strategy, Floating Material Suppression (FMS) loss and key physics features \textit{U}, \textit{PSL}, \textit{SED} for the \textbf{Test 1 (in-distribution)} and the \textbf{Test 2 (out-of-distribution)} dataset. 

\begin{table}[htbp]
    \centering
    \begin{tabular}{lcccccc}
        \toprule
        Metrics (\%) & \name{} & - HPG & - FMS & - U & - PSL & - SED \\
        \midrule
        Average CE $\downarrow$ & \textbf{0.87} & 4.34 & 1.82 & 4.21 & 2.79 & 3.89 \\
        Median CE $\downarrow$ & 0.16 & 0.48 & \textbf{0.13} & 0.41 & 0.24 & 0.33 \\
        Average VFE $\downarrow$ & \textbf{1.38} & 1.92 & 1.51 & 1.66 & 1.56 & 1.68 \\
        FM $\downarrow$ & \textbf{2.90} & 4.22 & 6.47 & 4.67 & 4.12 & 5.11 \\
        \bottomrule
    \end{tabular}
    \caption{Ablation study on key components for the \textbf{Test 1 (in-distribution)} dataset.}
    \label{tab:four_1}
\end{table}

\begin{table}[htbp]
    \centering
    \begin{tabular}{lcccccc}
        \toprule
        Metrics (\%) & \name{} & - HPG & - FMS & - U & - PSL & - SED \\
        \midrule
        Average CE $\downarrow$ & \textbf{5.29} & 36.62 & 7.74 & 18.14 & 11.92 & 15.31 \\
        Median CE $\downarrow$ & \textbf{0.61} & 1.26 & 0.88 & 1.05 & 0.68 & 0.74 \\
        Average VFE $\downarrow$ & \textbf{1.45} & 1.88 & 1.53 & 1.75 & 1.68 & 1.72 \\
        FM $\downarrow$ & \textbf{2.44} & 4.40 & 5.30 & 5.00 & 4.19 & 5.28 \\
        \bottomrule
    \end{tabular}
    \caption{Ablation study on key components for the \textbf{Test 2 (out-of-distribution)} dataset.}
    \label{tab:four_2}
\end{table}

\subsubsection{Ablation on hierarchical physics-guided strategy}
\revise{The "-HPG" variant in Table~\ref{tab:four_1} and Table~\ref{tab:four_2} keeps all three physics features, but removes the proposed hierarchical guidance. The three features are concatenated as a single non-hierarchical conditioning input and fed into the UNet without layer-specific cross-attention guidance.}
For in-distribution tests, the HPG strategy presents impressive enhancements across all metrics. Removing this strategy causes the Average CE and Median CE to increase from 0.87\% and 0.16\% to 4.34\% and 0.48\%, respectively. VFE and FM also increase. This performance degradation is even more pronounced in out-of-distribution tests, where the Average CE dramatically raises from 5.29\% to 36.62\%. This result demonstrates that the HPG is the keypoint of our method's generalizability to unseen conditions. Notably, the median CE is also less than 1\%, which means majority of generated structures are close to optimal. 

\subsubsection{Ablation on floating material suppression loss}

For in-distribution tests, incorporating FMS loss reduces FM from 6.47\% to 2.90\%. However, Average CE drops from 1.82\% to 0.87\% while the Median CE rises from 0.13\% to 0.16\%, which means adding FMS loss reduces the high value errors. We believe that this happens because the FMS loss explicitly guides the diffusion process towards generating designs with better structural connectivity. In TO, abnormally high compliance errors typically arise from disconnectivity. By penalizing invalid artifacts, the FMS loss effectively mitigates the instances causing these outlier errors, thus lowering the mean, while having less impact on the typical, well-connected cases reflected by the median. 

\begin{figure}[htbp] 
    \centering 
    \scalebox{0.7}{%
        \begin{minipage}[t]{0.48\columnwidth}
            \centering
            \textbf{Negative Samples}
    
            \begin{subfigure}[b]{0.31\linewidth}
                \centering
                \includegraphics[width=\linewidth]{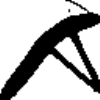}
                \label{fig:bad_12p_col3_sample_1}
            \end{subfigure}
            \hfill
            \begin{subfigure}[b]{0.31\linewidth}
                \centering
                \includegraphics[width=\linewidth]{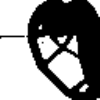}
                \label{fig:bad_12p_col3_sample_2}
            \end{subfigure}
            \hfill
            \begin{subfigure}[b]{0.31\linewidth}
                \centering
                \includegraphics[width=\linewidth]{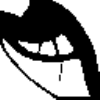}
                \label{fig:bad_12p_col3_sample_3}
            \end{subfigure}
    
            \begin{subfigure}[b]{0.31\linewidth}
                \centering
                \includegraphics[width=\linewidth]{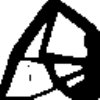}
                \label{fig:bad_12p_col3_sample_4}
            \end{subfigure}
            \hfill
            \begin{subfigure}[b]{0.31\linewidth}
                \centering
                \includegraphics[width=\linewidth]{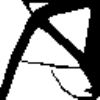}
                \label{fig:bad_12p_col3_sample_5}
            \end{subfigure}
            \hfill
            \begin{subfigure}[b]{0.31\linewidth}
                \centering
                \includegraphics[width=\linewidth]{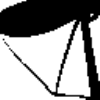}
                \label{fig:bad_12p_col3_sample_6}
            \end{subfigure}
    
            \begin{subfigure}[b]{0.31\linewidth}
                \centering
                \includegraphics[width=\linewidth]{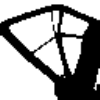}
                \label{fig:bad_12p_col3_sample_7}
            \end{subfigure}
            \hfill
            \begin{subfigure}[b]{0.31\linewidth}
                \centering
                \includegraphics[width=\linewidth]{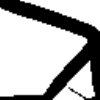}
                \label{fig:bad_12p_col3_sample_8}
            \end{subfigure}
            \hfill
            \begin{subfigure}[b]{0.31\linewidth}
                \centering
                \includegraphics[width=\linewidth]{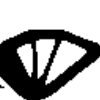}
                \label{fig:bad_12p_col3_sample_9}
            \end{subfigure}
    
            \begin{subfigure}[b]{0.31\linewidth}
                \centering
                \includegraphics[width=\linewidth]{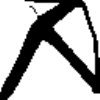}
                \label{fig:bad_12p_col3_sample_10}
            \end{subfigure}
            \hfill
            \begin{subfigure}[b]{0.31\linewidth}
                \centering
                \includegraphics[width=\linewidth]{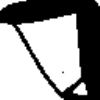}
                \label{fig:bad_12p_col3_sample_11}
            \end{subfigure}
            \hfill
            \begin{subfigure}[b]{0.31\linewidth}
                \centering
                \includegraphics[width=\linewidth]{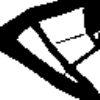}
                \label{fig:bad_12p_col3_sample_12}
            \end{subfigure}
        \end{minipage}
        \hfill 
        \begin{minipage}[t]{0.48\columnwidth}
            \centering
            \textbf{Positive Samples}
    
            \begin{subfigure}[b]{0.31\linewidth}
                \centering
                \includegraphics[width=\linewidth]{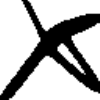}
                \label{fig:good_12p_col3_sample_1}
            \end{subfigure}
            \hfill
            \begin{subfigure}[b]{0.31\linewidth}
                \centering
                \includegraphics[width=\linewidth]{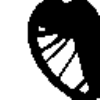}
                \label{fig:good_12p_col3_sample_2}
            \end{subfigure}
            \hfill
            \begin{subfigure}[b]{0.31\linewidth}
                \centering
                \includegraphics[width=\linewidth]{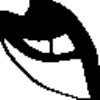}
                \label{fig:good_12p_col3_sample_3}
            \end{subfigure}
    
            \begin{subfigure}[b]{0.31\linewidth}
                \centering
                \includegraphics[width=\linewidth]{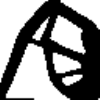}
                \label{fig:good_12p_col3_sample_4}
            \end{subfigure}
            \hfill
            \begin{subfigure}[b]{0.31\linewidth}
                \centering
                \includegraphics[width=\linewidth]{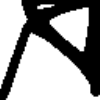}
                \label{fig:good_12p_col3_sample_5}
            \end{subfigure}
            \hfill
            \begin{subfigure}[b]{0.31\linewidth}
                \centering
                \includegraphics[width=\linewidth]{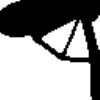}
                \label{fig:good_12p_col3_sample_6}
            \end{subfigure}
    
            \begin{subfigure}[b]{0.31\linewidth}
                \centering
                \includegraphics[width=\linewidth]{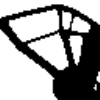}
                \label{fig:good_12p_col3_sample_7}
            \end{subfigure}
            \hfill
            \begin{subfigure}[b]{0.31\linewidth}
                \centering
                \includegraphics[width=\linewidth]{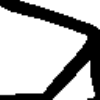}
                \label{fig:good_12p_col3_sample_8}
            \end{subfigure}
            \hfill
            \begin{subfigure}[b]{0.31\linewidth}
                \centering
                \includegraphics[width=\linewidth]{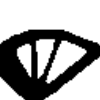}
                \label{fig:good_12p_col3_sample_9}
            \end{subfigure}
    
            \begin{subfigure}[b]{0.31\linewidth}
                \centering
                \includegraphics[width=\linewidth]{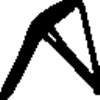}
                \label{fig:good_12p_col3_sample_10}
            \end{subfigure}
            \hfill
            \begin{subfigure}[b]{0.31\linewidth}
                \centering
                \includegraphics[width=\linewidth]{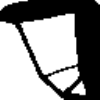}
                \label{fig:good_12p_col3_sample_11}
            \end{subfigure}
            \hfill
            \begin{subfigure}[b]{0.31\linewidth}
                \centering
                \includegraphics[width=\linewidth]{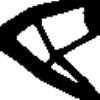}
                \label{fig:good_12p_col3_sample_12}
            \end{subfigure}
        \end{minipage}
    }
    \caption{Negative (w/o FMS loss) vs Positive (w/ FMS loss) samples in \textbf{out-of-distribution} tests. \revise{Negative samples contain visible floating material artifacts, while positive samples show the corresponding FMS-enabled results with improved material connectivity.}}
    \label{fig:four}
\end{figure}

For out-of-distribution tests, FMS loss improves all metrics, where Average CE decreases from 7.74\% to 5.29\%, Median CE drops from 0.88\% to 0.61\% and FM drops from 5.30\% to 2.44\%. This shows FMS loss can efficiently encourage the generation towards connected designs. Fig.~\ref{fig:four} presents examples for the comparison of generated results between \name{} and \name{} w/o FMS loss.

\subsubsection{Ablation on key physics features}

Furthermore, we investigate the individual contributions of the key physics features used in our guidance strategy: \textit{U}, \textit{PSL}, and \textit{SED}. Removing any of these three features leads to a noticeable degradation in performance across all metrics compared to full model.

Interestingly, the analysis reveals different contributions among these features, especially in the more challenging out-of-distribution tests. In \textbf{Test 2}, removing \textit{U} and \textit{SED} results in a larger performance drop than removing \textit{PSL}. For instance, the Average CE increases to 18.14\% without \textit{U} and 15.31\% without \textit{SED}, both of which are considerably higher than the 11.92\% when removing \textit{PSL}. A similar trend is observed in the other metrics. This suggests that while all three physics features are beneficial, \textit{U} and \textit{SED} appear to be more critical than \textit{PSL} for the model's generalizability.

\subsection{Case study}
\label{sec:case}

Most existing generative approaches for TO are image-based and thus inherently constrained to square design domains, while in practical engineering non-square domains are very common. For adapting to non-square domains, we employ parameter-efficient fine-tuning with LoRA~\cite{hu2022lora}. The success of this strategy is highly dependent on the generalizability of the pre-trained model. The pre-trained model must act as a strong generative prior, having learned the fundamental and transferable features of valid topologies from the initial square-domain data, which LoRA then specializes for problems with different size of design domains. To better validate the applicability of \name{} in more common scenarios, we conduct three case studies. 

\begin{figure}[htbp]
\centering
  \includegraphics[width=0.8\columnwidth]{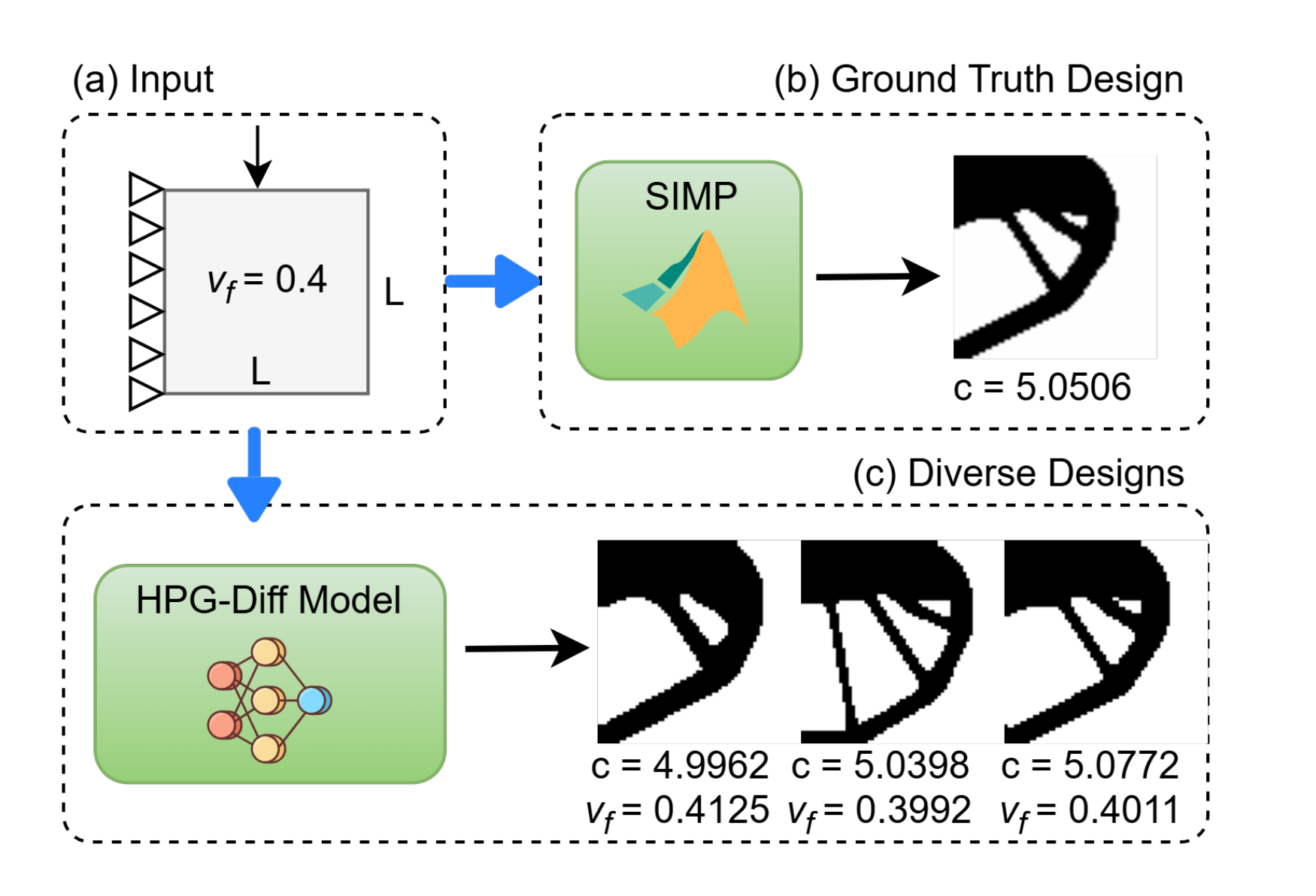} 
  \caption{Case study 1: Application of \name{} in shelf bracket design.}
  \label{fig:case_1}
\end{figure}

\textbf{Square Domain Study:} A \textbf{shelf bracket} is considered within a standard square design domain. 
Shelf brackets are common load-bearing components in furniture and mechanical assemblies.
This case serves as an illustrative baseline to demonstrate the performance of \name{} in generating diverse, high-performing solutions under a regular design domain. 
As shown in Fig.~\ref{fig:case_1}, the ground truth solution obtained by SIMP achieves a compliance of 5.0506. 
Our method produces three alternative layouts with compliance values of 4.9962, 5.0398, and 5.0772, which are all comparable to or better than the ground truth. 
Notably, these alternatives differ in structural characteristics—one with fewer voids, one with more distributed voids, and one with an intermediate configuration (featuring one, three, or two internal struts respectively). 

\textbf{Non-Square Domain Study:} To evaluate the adaptation potential of HPG-Diff on non-square rectangular domains, we construct a small dataset of 1,000 samples with domain size 60$\times$20 for fine-tuning the pretrained HPG-Diff using LoRA.  Two representative structures are investigated as case studies: the cantilever beam, which is widely used in mechanical and civil applications such as overhanging platforms, and the classic bridge, a typical TO benchmark relevant to civil engineering. In both cases, structural diversity enables engineers to compare alternatives for secondary performance criteria, such as fatigue life and vibration modes. 

\begin{figure}[htbp]
  \includegraphics[width=\columnwidth]{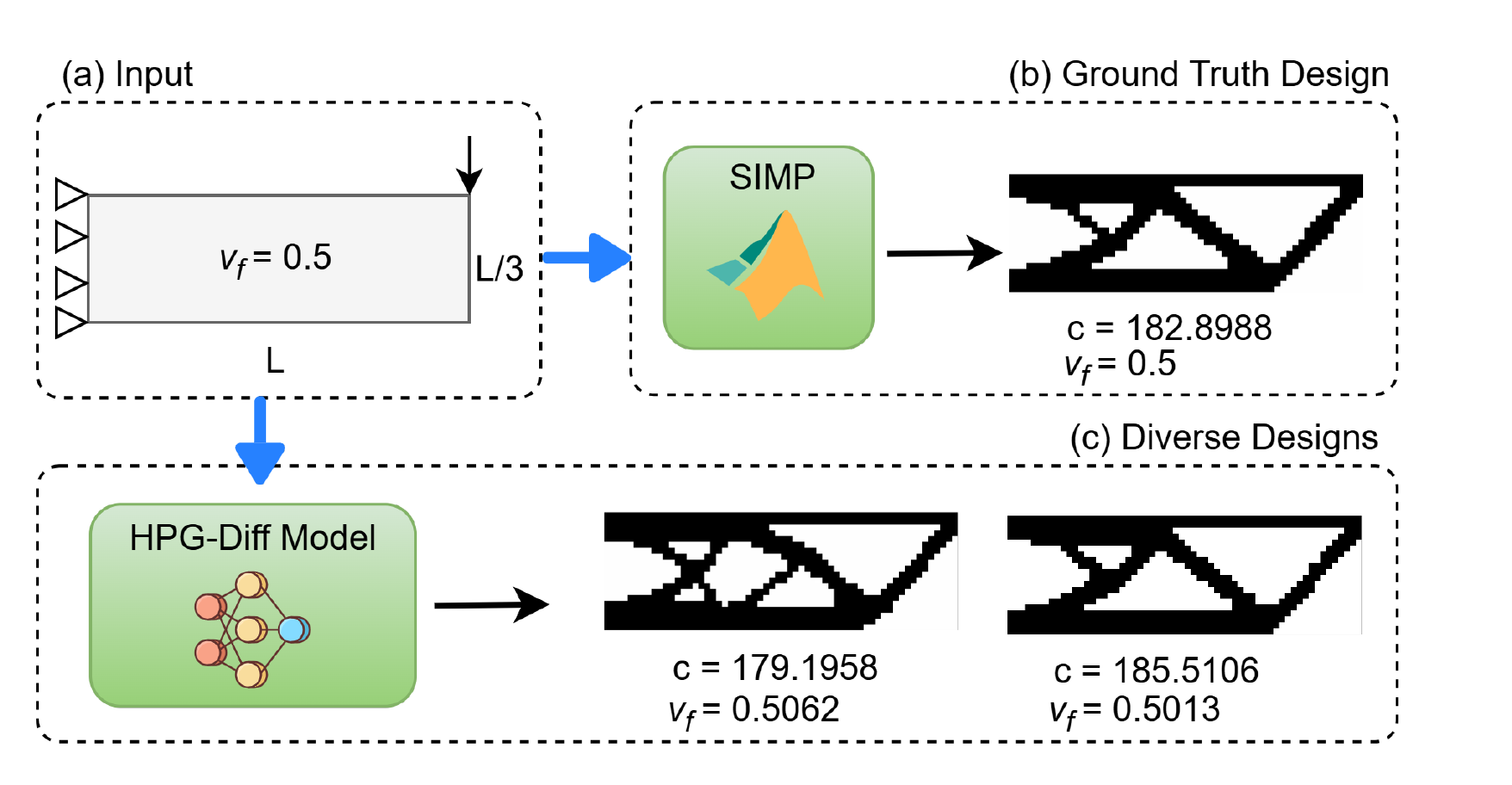} 
  \caption{Case study 2: Application of \name{} in cantilever beam design.}
  \label{fig:case_2}
\end{figure}

For cantilever beam (Fig.~\ref{fig:case_2}), the ground truth compliance is 182.8988. Our method generates two designs with compliance values of 179.1958 and 185.5106. The first alternative, with lower compliance, introduces more voids and supporting struts, which may enhance stiffness distribution but increase manufacturing complexity. The second design closely matches the ground truth in both compliance and geometry, providing a reliable baseline solution. Together, these alternatives allow engineers to balance lightweighting with ease of fabrication.

\begin{figure}[htbp]
  \includegraphics[width=\columnwidth]{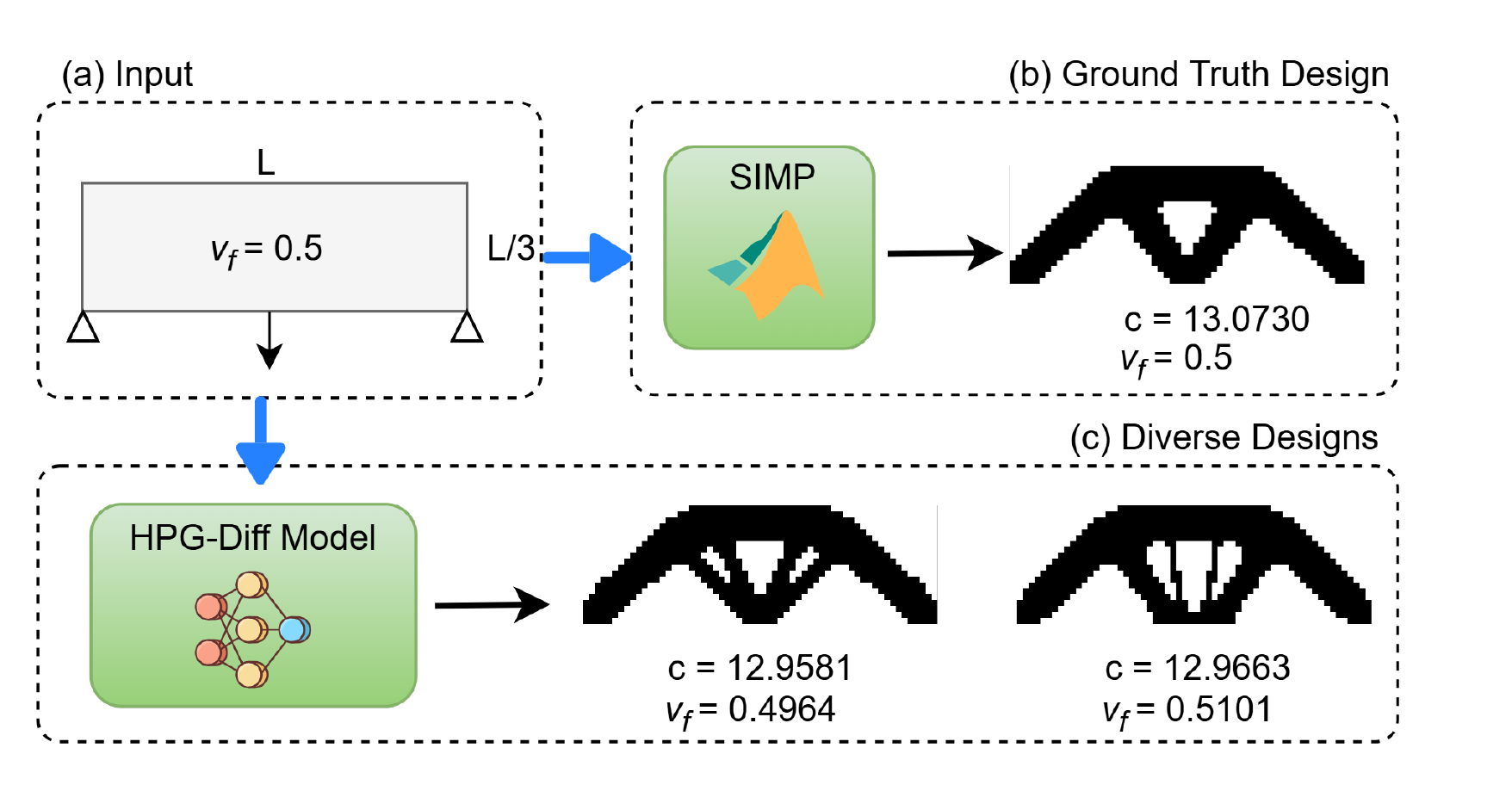} 
  \caption{Case study 3: Application of \name{} in classic bridge design.}
  \label{fig:case_3}
\end{figure}

For classic bridge (Fig.~\ref{fig:case_3}), the ground truth compliance is 13.0730, while our two designs achieve 12.9581 and 12.9663. Both alternatives slightly outperform the ground truth in compliance, but introduce more voids and internal struts. From an engineering perspective, these variants offer potential advantages in load distribution and redundancy, though they may require more intricate manufacturing processes.

\revise{To provide a quantitative evaluation in addition to the case studies, we additionally construct an independent 200-sample evaluation set with the same 60$\times$20 rectangular-domain and boundary conditions. These samples are not used during previous LoRA fine-tuning. We evaluate the LoRA-adapted HPG-Diff model on this independent set using CE, VFE, and FM.}

\begin{table}[h]
\centering
\small
\setlength{\tabcolsep}{6pt}
\caption{Quantitative evaluation of LoRA-adapted \name{} on an independent 200-sample 60$\times$20 rectangular-domain evaluation set.}
\label{tab:lora_rect_eval}
\begin{tabular}{cccc}
\toprule
Avg CE (\%) $\downarrow$ & Med CE (\%) $\downarrow$ & Avg VFE (\%) $\downarrow$ & FM (\%) $\downarrow$ \\
\midrule
2.86 & 0.58 & 1.54 & 3.17 \\
\bottomrule
\end{tabular}
\end{table}

\revise{As shown in Table~\ref{tab:lora_rect_eval}, the LoRA-adapted \name{} achieves an Avg CE of 2.86\%, a Med CE of 0.58\%, an Avg VFE of 1.54\%, and an FM ratio of 3.17\% on the independent evaluation set. These results quantitatively support the potential of lightweight LoRA fine-tuning for the 60$\times$20 rectangular-domain adaptation. The cantilever and bridge cases above further illustrate representative generated layouts under this adapted domain.}

\textbf{Findings:} Notably, several designs produced by \name{} outperform the SIMP baseline in compliance. We attribute this to two complementary factors: (1) stochastic sampling enables escape from local optima and thus broader exploration of the design space; (2) hierarchical physics guidance enforces basic topological correctness, keeping sampled solutions physically plausible and relatively accurate. As a result, the generated solutions form a variety of structural layouts (different void patterns and strut counts) that achieve comparable or better stiffness while providing designers with actionable alternatives across performance, manufacturability, and aesthetics. LoRA-based adaptation further verifies this when transferring to new domains.

\subsection{Discussion}

\subsubsection{Physic features orthogonality and redundancy}

To explore the rationality of chosen physics features, we analyze the orthogonality of these features to judge the redundancy. The primary metric for this analysis is the orthogonality score, calculated as $1 - |\cos(\theta)|$, where $\theta$ is the angle between the feature vectors. 
A score approaching 1 signifies high orthogonality, which represents strong independence. 

The orthogonality analysis was conducted on the training dataset. The results reveal varying degrees of orthogonality among the feature pairs. The highest degree of orthogonality was observed between \textit{SED} and \textit{PSL} (with a score of 0.8960±0.0527), followed by \textit{U} and \textit{SED} (0.7464±0.1048). In contrast, the \textit{U} and \textit{PSL} pair exhibited a significantly lower orthogonality score of 0.2065±0.1013. 
\revise{We also compute the orthogonality scores on Test 2 set. 
The same trend is observed: \textit{SED} and \textit{PSL} ($0.899 \pm 0.051$), \textit{U} and \textit{SED} ($0.741 \pm 0.107$), \textit{U} and \textit{PSL} ($0.202 \pm 0.100$).
This suggests that the redundancy pattern among the physics features is stable under the evaluated unseen boundary conditions.}
Statistically, this high correlation implies that \textit{PSL} contains information largely overlapping with \textit{U}.

\revise{However, this statistical view does not fully explain the asymmetrical importance revealed in the ablation study. 
Although \textit{U} and \textit{PSL} are correlated, they serve different roles in the hierarchical guidance process. 
\textit{U} provides a dense, boundary- and load-conditioned deformation response for low-level structural refinement, whereas \textit{PSL} explicitly represents internal load-transmission paths and provides a localized mid-level structural-skeleton guidance. 
This explains why removing \textit{U} causes the largest degradation, while removing \textit{PSL} still leads to a consistent but smaller performance drop.}

\revise{We further support this interpretation with feature-map visualizations in Appendix B. 
The \textit{PSL}-conditioned branch shows better relative concentration on the predicted X-shaped structural skeleton than the \textit{U}-conditioned branch, suggesting that \textit{PSL} does not merely duplicate \textit{U} but provides a more explicit mid-level skeleton cue.}

\begin{table}[htbp]
    \centering
    \setlength{\tabcolsep}{2pt}
    \begin{tabular}{@{}lccccc@{}}
        \toprule
        \textbf{Method} & \textbf{Preprocess} & \textbf{Sampling} & \textbf{Postprocess}  & \textbf{Inference} & \textbf{$\Delta{T}_{inf}$} \\
        \midrule
        SIMP & - & - & 23.93 & 23.93 & 0.00 \\
        TopoDiff & 3.83 & 3.12 & - & 6.95 & - 70.96 \\
        TopoDiff-Guided & 3.83 & 22.65 & - & 26.48 & + 10.66 \\
        DOM w/ TA & 0.12 & 3.03 & - & 3.15 & - 86.84 \\
        \textbf{\name{}} & 5.92 & 4.82 & - & 10.74 & - 55.12 \\
        \bottomrule
    \end{tabular}
    \caption{Inference time (s) comparison for different methods (64x64 domain, 100 steps). All methods were tested on a single GPU, except the SIMP on a multi-core CPU. $\Delta{T}_{inf}$ (\%) is the relative time difference compared with SIMP.}
    \label{tab:six}
\end{table}

\subsubsection{Inference efficiency and generalizability}

A key distinction of our work lies in the simplicity during the inference phase and effectiveness of the guidance mechanism during training phase. Deep learning methods try to avoid costly simulations within the inference phase. Specifically, guided-diffusion approaches like TopoDiff employ auxiliary models trained with FEA generated physics features to guide the inference phase. DOM refines this paradigm by substituting the expensive FEA preprocessing features with a fast kernel relaxation, and employing trajectory alignment that uses intermediate optimization results to guide the model's training process.

While this approach eliminates guidance in the inference phase and reduces inference time, it results in a notable trade-off in performance. This strategy for reducing inference cost is instructive. However, the large degradation in performance highlights the critical importance of effective physics guidance in constrained generation tasks.

\revise{In contrast, our approach computes initial key physics features via FEA. All finite elements are assigned the same initial density, and a single FEA is solved under the prescribed boundary conditions and loads. Therefore, the physics features are computed from a uniform density field rather than from the optimized target topology. For \name{}, the preprocessing time denotes one FEA solve together with the subsequent feature calculation. 
While TopoDiff and TopoDiff-Guided also involve FEA-based preprocessing to obtain the physics fields used for conditioning, DOM w/ TA replaces such FEA-based preparation with a lightweight kernel relaxation step. Therefore, the preprocessing column in Table~\ref{tab:six} denotes the method-specific preparation required before diffusion sampling: FEA-based physics-feature preparation for TopoDiff/TopoDiff-Guided, kernel relaxation for DOM w/ TA, and one FEA solve plus feature calculation for \name{}. This clarification ensures that the reported inference time reflects the full pipeline cost of each method rather than only the diffusion sampling time.}
As detailed in Table~\ref{tab:six}, our method \name{} clocks in at 10.74 seconds. While not as fast as DOM, our method is approximately twice as fast as the classical SIMP method and outperforms methods like TopoDiff-Guided and DOM. This demonstrates that our "preprocess and hierarchically guide" paradigm strikes a more effective balance. It achieves better accuracy and generalizability with a practical inference time, avoiding both the performance trade-offs of overly simplified models and the computational cost of inference phase guidance.

\revise{The out-of-distribution results support the generalization potential of HPG-Diff within the evaluated benchmark, particularly for unseen boundary conditions. Nevertheless, this claim remains limited under constraints of our experimental setup, and broader generalizability requires further validation across more diverse settings.}
However, we believe that the core approach of our hierarchical physics-guided diffusion is broadly applicable to other physics-based design problems. Indeed, diffusion-based generative frameworks have recently been successfully applied to complex challenges in other fields, such as fluid dynamics~\cite{luo2024difffluid, qiu2024pi} and aerodynamic optimization~\cite{deng2025generative}, suggesting the potential for adapting our specific guidance strategy to these domains as well.

While we currently ignore certain advanced physics properties (such as non-linear material responses, fatigue behavior, or dynamic loading conditions) or manufacturing constraints for 3D printing (such as minimum and maximum length scales, overhang limitations, or surface roughness), our method takes a step towards precisely incorporating physics features that have implications for performance and manufacturability into data-driven methods. Aiming for enhanced physics consistency and more readily deployable designs in real-world scenarios, we seek to include a broader and more specific set of physics properties and manufacturing constraints with minimal additional costs in the future. This could involve designing novel loss functions that directly penalize violations of these specific constraints like our FMS loss, or exploring architectural modifications to better capture multi-scale physics principles.

\subsubsection{Design diversity and engineering applicability}
Besides efficiency and performance, design diversity and engineering applicability are equally critical criteria. 
As shown in Section~\ref{sec:case}, \name{} successfully generates diverse, high-performing solutions in both square and $3{:}1$ rectangular domains. 
Design diversity is particularly important in practice, as multiple alternatives allow engineers to balance performance with other factors such as manufacturability, cost, or even aesthetic considerations.

Nevertheless, the current engineering validation remains bounded in two aspects. First, the domain-adaptation study is limited to relatively simple aspect-ratio changes. \revise{The quantitative benchmark follows the fixed $64\times64$ resolution of the TopoDiff~\cite{maze2023diffusion} dataset. A multi-resolution study would require rebuilding the dataset rather than simply resizing the outputs, since $128\times128$ and $256\times256$ domains contain 4 and 16 times more finite elements per sample, respectively.} More complex scenarios, such as L-shaped connectors, circular domains (e.g., bearing housings or pressure vessels), and fully irregular shapes (common in biomedical implants or customized mechanical parts), have not yet been addressed. While LoRA-based fine-tuning shows adaptability for rectangular domains, its success relies heavily on task similarity. \revise{LoRA adaptation also still relies on target-domain samples generated by conventional TO solvers. Therefore, this case study is intended to examine the convenience of lightweight adaptation from existing SIMP-generated data to AI-based rapid generation, rather than to remove the cost of obtaining data. Future work should investigate more robust and data-efficient adaptation strategies, such as incorporating multi-shape domains in the training dataset, proposing stronger geometry-aware conditioning, or reducing the cost of acquiring reliable training data.}

\revise{Second, the current benchmark contains single-load cases, and the FMS loss is therefore instantiated with heat propagation initialized from the prescribed load region. This design follows the role of FMS in our framework: the main load-bearing material distribution is guided by the boundary-conditioned physics features in HPG, which encourage the formation of a load-transfer structure between the applied load and the constrained supports. However, the generated density field may still contain noisy floating material artifacts, such as disconnected islands or weakly connected fragments detached from the main load-transfer body. FMS mitigates this by propagating from the load region through the generated material layout to support regions and penalizing material that is not effectively incorporated into this load-transfer body. For problems with multiple prescribed loads, the seed mask can be extended as the union of all load regions. Future work should evaluate such multi-load settings.}

\revise{Besides these limitations, uncertainty quantification is also an important future direction for stochastic generative TO. 
Although the repeated-sampling analysis in Appendix D provides an empirical view of output variability across inference seeds, it does not constitute formal uncertainty quantification. 
Future work could combine generative TO with Bayesian uncertainty modeling~\cite{bhattacharya2026optimal,mahmoud2026sequential}.}

\section{Conclusion}

\revise{In this paper, we propose \name{}, a novel diffusion framework for topology optimization.
By synergizing a hierarchical physics-guided strategy with a differentiable connectivity constraint, HPG-Diff improves generation quality under unseen boundary conditions and reduces floating material artifacts in generated density fields.
Numerical results demonstrate that aligning key physics features with the denoising process helps capture structural mechanics, reducing the average compliance error to 5.29\% in out-of-distribution tests and reducing the floating material ratio to 2.44\%.
Consequently, \name{} achieves the best performance compared with representative generative TO baselines, without requiring external models or post-processing. The case studies further indicate the potential of lightweight LoRA fine-tuning for the adaptation of HPG-Diff to rectangular non-square domains.
Overall, HPG-Diff provides an effective framework for integrating physics guidance into generative topology optimization.}

\vspace{-0.5em}

\section*{Data availability}
Data will be made available on request.

\vspace{-2em}

\revise{\section*{Funding}}
\revise{This research was supported by the National Key R\&D Program of China (No. 2023YFB4604800).}







\appendix
\revise{\section{Additional Ablations on \name{}}}
\label{app:additional_hpg_ablation}

\revise{To further clarify the contribution of the hierarchical physics-guided strategy, we conduct two additional ablation studies.
First, "Bottleneck" concatenates U, PSL, and SED and injects them only at the bottleneck layer as a single conditioning signal. This variant tests whether a single global injection of all physics features is sufficient without layer-specific assignment. 
Second, U-SED Swap variant keeps the same physics features and the same layer-specific cross-attention structure, but swaps the shallow-layer U guidance and the deep-layer SED guidance. This variant tests whether the feature-level matching is important while keeping other settings unchanged.}

\revise{As shown in Table~\ref{tab:app_hpg_ablation}, \name{} reduces Average CE from 3.70\% and 2.43\% to 0.87\%, and reduces FM from 3.79\% and 4.10\% to 2.90\%, compared with Bottleneck and U-SED Swap in Test 1, respectively. On the more challenging Test 2, the improvement is more pronounced: \name{} reduces Average CE from 16.42\% and 13.18\% to 5.29\%, and reduces FM from 4.18\% and 4.61\% to 2.44\%. The results suggest that the performance gain is not merely due to providing additional physics features, but also to organizing these features according to the proposed hierarchical physics-guidance strategy.}

\begin{table}[h]
\centering
\setlength{\tabcolsep}{4pt}
\caption{
Additional ablations on hierarchical physics guidance. 
\textbf{Bottleneck} concatenates \textit{U}, \textit{PSL}, and \textit{SED} and injects them only at the bottleneck layer.
\textbf{U-SED Swap} keeps the layer-specific cross-attention structure but exchanges the shallow-layer \textit{U} guidance and deep-layer SED guidance.
}
\label{tab:app_hpg_ablation}
\begin{tabular}{llcccc}
\toprule
Dataset & Variant & Average CE$\downarrow$ & Median CE$\downarrow$ & Average VFE$\downarrow$ & FM$\downarrow$ \\
\midrule
\multirow{3}{*}{Test 1}
& \name{} & \textbf{0.87} & \textbf{0.16} & \textbf{1.38} & \textbf{2.90} \\
& Bottleneck & 3.70 & 0.35 & 1.68 & 3.79 \\
& U-SED Swap & 2.43 & 0.27 & 1.58 & 4.10 \\
\midrule
\multirow{3}{*}{Test 2}
& \name{} & \textbf{5.29} & \textbf{0.61} & \textbf{1.45} & \textbf{2.44} \\
& Bottleneck & 16.42 & 0.97 & 1.73 & 4.18 \\
& U-SED Swap & 13.18 & 0.82 & 1.65 & 4.61 \\
\bottomrule
\end{tabular}
\end{table}

\revise{\section{Feature Map Visualizations of \textit{U} and \textit{PSL}}}
\label{app:psl_feature_map}

\revise{To address the concern that \textit{PSL} may be redundant with \textit{U} due to their high correlation, we further examine whether \textit{PSL} contributes a distinct representational role in the denoising process. 
Specifically, we visualize the cross-attention output feature maps conditioned on \textit{U} and \textit{PSL} during late denoising.  
Let $O_U$ and $O_{PSL}$ denote the output tensors of the corresponding cross-attention modules. 
For each timestep, we aggregate the output tensor over channels using the mean absolute value, and visualize the resulting feature map at the native resolution.}

\revise{We use the same example as in Fig.~\ref{fig:two} for consistency with the previous sections, enabling a direct visual comparison among the conditioning signals, the cross-attention output feature maps, and the predicted topology. 
We focus on late denoising timesteps ($t=50$ to $t=0$, with $T=1000$), since early-stage predictions are dominated by noise and do not yet provide a stable structural layout for visual comparison.}

\begin{figure}[t]
\centering
\includegraphics[width=\linewidth]{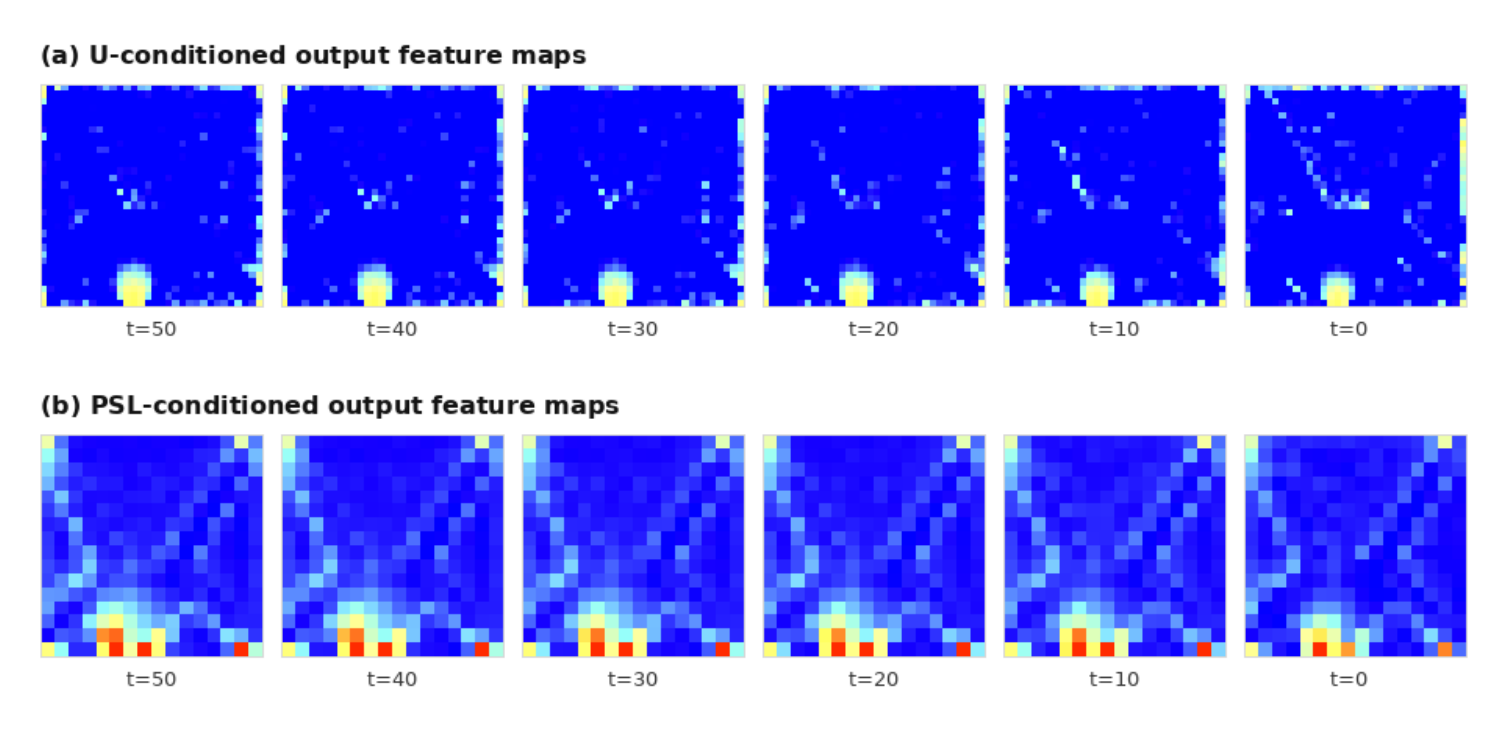}
\caption{
Feature-map visualization of cross-attention outputs. 
The upper row shows \textit{U}-conditioned cross-attention output feature maps, and the lower row shows \textit{PSL}-conditioned cross-attention output feature maps. 
}
\label{fig:psl_activation_vis}
\end{figure}

\revise{As shown in Fig.~\ref{fig:psl_activation_vis}, the \textit{PSL}-conditioned output feature maps exhibit more concentrated responses along the X-shaped structural skeleton and the central load-transfer region. 
In contrast, the \textit{U}-conditioned feature maps are more spatially diffuse. 
We further quantify this observation in Table~\ref{tab:psl_activation_concentration} by using the predicted X-shaped topology region as a mask. 
The \textit{PSL}-conditioned result consistently achieves an $M_X/M_{\mathrm{bg}}$ ratio larger than 1.0 across late denoising timesteps, while the \textit{U}-conditioned result remains below 1.0. 
These results suggest that \textit{PSL} does not merely duplicate the \textit{U}-conditioned response; instead, it provides a more explicit mid-level structural-skeleton cue for the denoising process.}

\revise{This observation is also consistent with the different physical roles of the two features in topology optimization. 
The displacement field \textit{U} provides a dense, boundary- and load-conditioned deformation response, indicating how the domain tends to deform under the prescribed problem setting. 
Such information is useful for low-level structural refinement, but it does not directly define the final load-bearing skeleton. 
In contrast, \textit{PSL} explicitly represents internal load-transmission paths, which are more closely aligned with the structural skeleton formed by the optimized topology.}

\begin{table}[t]
\centering
\small
\setlength{\tabcolsep}{5pt}
\caption{
Concentration of cross-attention output feature responses on the predicted topology region.
$M_X$ denotes the mean feature response inside the predicted topology mask, where the response map is computed as $M(i,j)=\mathrm{mean}_c(|O_c(i,j)|)$ from the corresponding cross-attention output tensor.
$M_X/M_{\mathrm{bg}}$ denotes the ratio between the masked-region response and the background response, measuring the relative concentration on the structural skeleton.
}
\label{tab:psl_activation_concentration}
\begin{tabular}{c|cc|cc}
\toprule
\multirow{2}{*}{Timestep} 
& \multicolumn{2}{c|}{Mean response $M_X$} 
& \multicolumn{2}{c}{Relative contrast $M_X/M_{\mathrm{bg}}$} \\
\cmidrule(lr){2-3} \cmidrule(lr){4-5}
& \textit{U}-conditioned & \textit{PSL}-conditioned 
& \textit{U}-conditioned & \textit{PSL}-conditioned \\
\midrule
50 & 0.215 & 0.396 & 0.777 & \textbf{1.065} \\
40 & 0.214 & 0.395 & 0.785 & \textbf{1.063} \\
30 & 0.213 & 0.394 & 0.776 & \textbf{1.060} \\
20 & 0.213 & 0.392 & 0.783 & \textbf{1.057} \\
10 & 0.211 & 0.390 & 0.770 & \textbf{1.049} \\
0  & 0.207 & 0.380 & 0.739 & \textbf{1.032} \\
\bottomrule
\end{tabular}
\end{table}

\revise{\section{Sensitivity Analysis of FMS Loss}}
\label{app:fms_sensitivity}

\revise{To further examine the proposed Floating Material Suppression (FMS) loss, we conduct sensitivity analyses on three parameters: the propagation step $k$, the time-weighting scalar $c$, and the FMS loss weight $\lambda_{\mathrm{fm}}$. These parameters play different roles in FMS. The propagation step $k$ controls the spatial coverage of the differentiable propagation process. The scalar $c$ controls when the FMS regularization becomes effective during DDPM denoising process through $w(t)=\exp(-ct/T)$, where $T$ is the number of diffusion timesteps. The weight $\lambda_{\mathrm{fm}}$ balances the standard denoising objective and the connectivity regularization. Therefore, we vary one parameter at a time while keeping the others default in our experiments ($k=64$, $T=1000$, $c=5$, $\lambda_{\mathrm{fm}}=0.1$).}

\revise{\subsection{Effect of propagation step $k$}
The parameter $k$ is selected according to the coverage requirement of the propagation process. Since the local propagation in Eq.~(6) expands the reachable region step by step, $k$ should be large enough to cover the design domain; otherwise, connected material far from the load seed may be incorrectly treated as unreachable. As shown in Table~\ref{tab:app_k_sensitivity}, a smaller $k$ weakens floating-material suppression, while increasing $k$ beyond the domain-covering range brings limited additional benefit. We therefore set $k=64$ for the $64\times64$ design domain.}

\begin{table}[h]
\centering
\small
\setlength{\tabcolsep}{5pt}
\caption{Sensitivity analysis of the propagation step $k$ in the FMS loss on Test 2. Other parameters are fixed at $c=5$ and $\lambda_{\mathrm{fm}}=0.1$.}
\label{tab:app_k_sensitivity}
\begin{tabular}{lcccc}
\toprule
$k$ & Avg CE (\%) $\downarrow$ & Med CE (\%) $\downarrow$ & Avg VFE (\%) $\downarrow$ & FM (\%) $\downarrow$ \\
\midrule
32 & 6.46 & 0.75 & 1.52 & 3.12 \\
64 & 5.29 & 0.61 & 1.45 & 2.44 \\
96 & 5.33 & 0.62 & 1.46 & 2.39 \\
\bottomrule
\end{tabular}
\end{table}

\revise{\subsection{Effect of time-weighting scalar $c$}}
\revise{The scalar $c$ controls the temporal schedule of the FMS loss. In the reverse DDPM process, larger $t$ corresponds to noisier states, while late reverse denoising corresponds to smaller, low-noise timesteps near zero. With $T=1000$ and the default $c=5$, the weight $w(t)=\exp(-5t/1000)$ is suppressed at high-noise timesteps, e.g., $w(1000)=0.0067$ and $w(500)=0.082$, and becomes much stronger at low-noise timesteps, e.g., $w(100)=0.607$ and $w(0)=1$. This is consistent with the role of FMS: penalizing disconnected components is more meaningful after the predicted material distribution has formed a relatively clear structure. As shown in Table~\ref{tab:app_c_sensitivity}, the default setting achieves the best CE--FM balance. Removing the temporal decay ($c=0$) applies the connectivity penalty too broadly across noisy denoising steps, while a larger value ($c=10$) delays the regularization excessively and weakens floating-material suppression.}

\begin{table}[h]
\centering
\small
\setlength{\tabcolsep}{6pt}
\caption{Sensitivity analysis of the time-weighting scalar $c$ in the FMS loss on Test 2. Other parameters are fixed at $k=64$ and $\lambda_{\mathrm{fm}}=0.1$.}
\label{tab:app_c_sensitivity}
\begin{tabular}{lcccc}
\toprule
$c$ & Avg CE (\%) $\downarrow$ & Med CE (\%) $\downarrow$ & Avg VFE (\%) $\downarrow$ & FM (\%) $\downarrow$ \\
\midrule
0   & 6.38 & 0.79 & 1.69 & 5.03 \\
2.5 & 5.71 & 0.67 & 1.52 & 2.82 \\
5   & 5.29 & 0.61 & 1.45 & 2.44 \\
10  & 5.62 & 0.66 & 1.49 & 3.11 \\
\bottomrule
\end{tabular}
\end{table}

\revise{\subsection{Effect of FMS weight $\lambda_{\mathrm{fm}}$}}

\revise{The weight $\lambda_{\mathrm{fm}}$ controls the relative strength of the connectivity regularization. Its role is not to override the main denoising objective, but to suppress disconnected high-density regions while preserving low compliance. As shown in Table~\ref{tab:app_lambda_sensitivity}, setting $\lambda_{\mathrm{fm}}=0$ removes the FMS regularization and leads to a higher FM ratio. Increasing $\lambda_{\mathrm{fm}}$ reduces floating material, but an overly large weight starts to degrade compliance and volume-fraction accuracy, indicating the expected trade-off between connectivity regularization and structural optimality. The default value $\lambda_{\mathrm{fm}}=0.1$ lies near the knee of this CE--FM trade-off: it reduces FM while achieving the lowest Avg CE among the tested settings.}

\begin{table}[h]
\centering
\small
\setlength{\tabcolsep}{6pt}
\caption{Sensitivity analysis of the FMS loss weight $\lambda_{\mathrm{fm}}$ on Test 2. Other parameters are fixed at $k=64$ and $c=5$.}
\label{tab:app_lambda_sensitivity}
\begin{tabular}{lcccc}
\toprule
$\lambda_{\mathrm{fm}}$ & Avg CE (\%) $\downarrow$ & Med CE (\%) $\downarrow$ & Avg VFE (\%) $\downarrow$ & FM (\%) $\downarrow$ \\
\midrule
0    & 7.74 & 0.88 & 1.53 & 5.30 \\
0.05 & 6.12 & 0.71 & 1.49 & 3.36 \\
0.10 & 5.29 & 0.61 & 1.45 & 2.44 \\
0.20 & 5.97 & 0.69 & 1.61 & 2.41 \\
0.40 & 7.08 & 0.93 & 1.80 & 2.18 \\
\bottomrule
\end{tabular}
\end{table}

\revise{Overall, these results show that the FMS loss remains stable under moderate parameter changes around the default setting, while extreme settings exhibit the expected trade-off between connectivity regularization and compliance accuracy.}

\revise{\section{Repeated Sampling across Inference Seeds}}
\label{app:multi_seed}

\revise{Since generative models are stochastic, we further evaluate \name{} with repeated inference runs across multiple random seeds. We use the same trained checkpoint and inputs, and only vary the inference-time random seed.}

\begin{table}[h]
\centering
\small
\setlength{\tabcolsep}{5pt}
\caption{Repeated inference results of \name{} over five stochastic inference trials.}
\label{tab:app_multiseed}
\begin{tabular}{lllc}
\toprule
Set & Metrics & Repeated results & Mean $\pm$ Std \\
\midrule
\multirow{4}{*}{Test 1}
& Avg CE (\%) $\downarrow$  & 0.87, 0.93, 0.97, 1.01, 0.81 & 0.92$\pm$0.08 \\
& Med CE (\%) $\downarrow$  & 0.16, 0.17, 0.18, 0.16, 0.17 & 0.17$\pm$0.01 \\
& Avg VFE (\%) $\downarrow$ & 1.38, 1.42, 1.47, 1.48, 1.47 & 1.44$\pm$0.04 \\
& FM (\%) $\downarrow$      & 2.90, 3.50, 3.12, 2.70, 3.33 & 3.11$\pm$0.32 \\
\midrule
\multirow{4}{*}{Test 2}
& Avg CE (\%) $\downarrow$  & 5.29, 5.58, 5.51, 8.32, 7.24 & 6.39$\pm$1.33 \\
& Med CE (\%) $\downarrow$  & 0.61, 0.58, 0.63, 0.58, 0.58 & 0.60$\pm$0.02 \\
& Avg VFE (\%) $\downarrow$ & 1.45, 1.45, 1.48, 1.42, 1.38 & 1.44$\pm$0.04 \\
& FM (\%) $\downarrow$      & 2.44, 3.40, 3.50, 2.32, 3.13 & 2.96$\pm$0.55 \\
\bottomrule
\end{tabular}
\end{table}

\revise{As shown in Table~\ref{tab:app_multiseed}, \name{} shows low variance on Test 1. On the more challenging Test 2, Med CE and Avg VFE remain stable across seeds while the standard deviation of Avg CE is 1.33\%. This result indicates that most samples remain stable across inference seeds, but the average compliance error can be affected by difficult long-tail cases.}

\revise{\section{Tail-Distribution and Failure-Case Analysis}}
\label{app:failure_cases}

\revise{We analyze the tail distribution of compliance error (CE) and inspect representative failure cases.
Following the high-error threshold used in TopoDiff~\cite{maze2023diffusion}, we define failure cases as samples with CE larger than 30\%. Table~\ref{tab:app_tail_stats} reports the distributional statistics of per-sample CE on Test 2 for the reference inference run reported in the main benchmark tables.}

\begin{table}[h]
\centering
\scriptsize
\setlength{\tabcolsep}{4pt}
\caption{Tail-distribution statistics of \name{} on Test 2 for the reference inference run reported in the main benchmark tables.}
\label{tab:app_tail_stats}
\begin{tabular}{lccccc}
\toprule
Method & Avg CE (\%) $\downarrow$ & Mdn CE (\%) $\downarrow$ & Std CE (\%) $\downarrow$ & Max CE (\%) $\downarrow$ & CE $>$ 30\% (\%) $\downarrow$ \\
\midrule
HPG-Diff & 5.29 & 0.61 & 31.76 & 791.07 & 2.70\% \\
\bottomrule
\end{tabular}
\end{table}

\revise{As shown in Table~\ref{tab:app_tail_stats}, the error distribution is long-tailed. The median CE is 0.61\%, and only 2.70\% of samples exceed the 30\% high-error threshold. However, the maximum CE reaches 791.07\%, indicating that Avg CE is affected by some severe failure cases. This explains the variance of Avg CE.}

\begin{figure}[h]
\centering
\includegraphics[width=\linewidth]{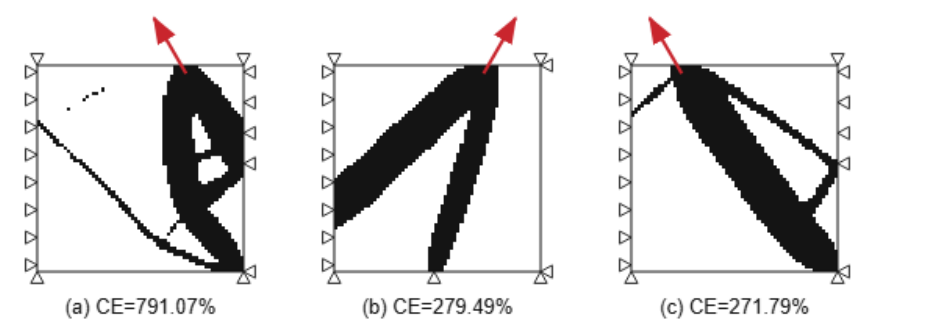}
\caption{Representative high-error OOD cases with boundary conditions overlaid. Hollow triangular markers denote constrained boundary supports, and red arrows denote applied loads.}
\label{fig:app_failure_cases}
\end{figure}

\revise{Fig.~\ref{fig:app_failure_cases} shows representative high-error cases with the corresponding boundary conditions overlaid. In these examples, high CE is associated with weak or incomplete load-transfer paths between the applied load and the constrained boundaries. Some generated topologies contain overly thin members or disconnected material, leading to insufficient stiffness.}

\revise{\section{Construction of Principal Stress Lines}}
\label{app:psl_construction}

\revise{The principal stress line (PSL) maps are constructed from the finite-element stress field following the standard principal-stress-line tracing procedure used in structural topology design~\cite{kwok2016structural}. For each element or grid point in the design domain, the in-plane stress tensor is written as
\begin{equation}
\boldsymbol{\sigma} =
\begin{bmatrix}
\sigma_x & \tau_{xy} \\
\tau_{xy} & \sigma_y
\end{bmatrix}.
\end{equation}
The two principal stresses and their corresponding directions are obtained by eigenvalue decomposition:
\begin{equation}
\boldsymbol{\sigma}\mathbf{v}_i = \sigma_i \mathbf{v}_i, \quad i=1,2,
\end{equation}
where $\sigma_1$ and $\sigma_2$ denote the maximum and minimum principal stresses, respectively, and $\mathbf{v}_1$ and $\mathbf{v}_2$ are the corresponding principal directions.}

\revise{Starting from uniformly sampled seed points in the solid design domain, PSLs are traced by integrating along the local principal stress directions. Tensile PSLs are traced along the maximum principal-stress direction $\mathbf{v}_1$, while compressive PSLs are traced along the minimum principal-stress direction $\mathbf{v}_2$. Line tracing is terminated when the trajectory reaches the domain boundary, enters a void region, or reaches a region with negligible principal stress magnitude.}

\revise{The traced PSLs are then rasterized onto the same spatial resolution as the model input. In this work, tensile and compressive PSLs are represented as two separate feature channels to avoid ambiguity between line type and line intensity. The rasterized maps are normalized before being used as physics-guided input features.}

\revise{\section{Additional Benchmark Comparisons}}
\label{app:additional_benchmark_comparison}

\revise{We further compare HPG-Diff with NITO~\cite{nobari2025nito} and OAT~\cite{nobari2025oat}.
NITO is a neural implicit field-based topology optimization model that represents boundary conditions with sparse point-based inputs.
OAT is positioned as a foundation model for structural topology optimization and is trained on OpenTO, a large-scale dataset containing 2.2 million optimized structures and 2 million unique boundary-condition configurations.}

\revise{Since OAT is positioned as a foundation model trained on OpenTO, we report this comparison on the in-distribution Test 1 set.
Retraining OAT only on our training data would not reflect its intended foundation-model setting, while directly comparing it on our Test 2 set would mix differences in training-data scope with the out-of-distribution evaluation.
Therefore, Test 1 provides a more suitable reference for comparing generation quality under the same benchmark distribution.}

\begin{table}[h]
\centering
\small
\setlength{\tabcolsep}{8pt}
\caption{Additional benchmark comparison on in-distribution Test 1.}
\label{tab:app_nito_oat}
\begin{tabular}{lccc}
\toprule
Method & Avg CE (\%) $\downarrow$ & Med CE (\%) $\downarrow$ & Avg VFE (\%) $\downarrow$ \\
\midrule
NITO~\cite{nobari2025nito} & 8.13 & 0.47 & 1.40 \\
OAT~\cite{nobari2025oat} & 1.74 & 0.32 & \textbf{0.25} \\
HPG-Diff (Ours) & \textbf{0.87} & \textbf{0.16} & 1.38 \\
\bottomrule
\end{tabular}
\end{table}

\revise{As shown in Table~\ref{tab:app_nito_oat}, HPG-Diff achieves the lowest compliance errors on the in-distribution Test 1, with an Avg CE of 0.87\% and a Med CE of 0.16\%, compared with 8.13\%/0.47\% for NITO and 1.74\%/0.32\% for OAT. This suggests that the hierarchical physics-guided conditioning and FMS help the model generate layouts that better match the load-transfer structure required for low compliance. In contrast, NITO uses a neural implicit representation with sparse boundary-condition inputs, which provides flexibility but may be less specialized for this benchmark. OAT, as a foundation model trained on a much broader OpenTO distribution, achieves the lowest Avg VFE of 0.25\%, indicating better volume fraction preservation. However, its broader training objective and cross-domain generalization setting may make it less specialized for minimizing compliance error on this specific dataset than HPG-Diff.}

\bibliographystyle{elsarticle-num-names}
\bibliography{cas-refs}

\end{document}